\documentclass{article}

    \PassOptionsToPackage{numbers, compress}{natbib}

\usepackage[preprint]{neurips_2026}

\usepackage[utf8]{inputenc} 
\usepackage[T1]{fontenc}    
\usepackage{hyperref}       
\usepackage{url}            
\usepackage{booktabs}       
\usepackage{amsfonts}       
\usepackage{nicefrac}       
\usepackage{microtype}      
\usepackage{xcolor}         
\usepackage{wrapfig}
\usepackage{amsmath} 
\usepackage{multirow}
\usepackage{microtype}
\usepackage{graphicx}
\usepackage{subfigure}
\usepackage{booktabs} 
\usepackage[ruled,vlined,linesnumbered]{algorithm2e}
\usepackage{colortbl}

\usepackage[utf8]{inputenc} 
\usepackage[T1]{fontenc}    
\usepackage{hyperref}       
\usepackage{url}            
\usepackage{booktabs}       
\usepackage{amsfonts}       
\usepackage{nicefrac}       
\usepackage{microtype}      
\usepackage{xcolor}         
\usepackage{colortbl}
\definecolor{customcolor}{HTML}{D3D3D3}
\usepackage{algorithmic}

\makeatletter
\renewcommand{\@notice}{}
\makeatother

\title{Think Densely, Act Sparsely: \\ Latent Expert Cognitive Chains for \\ Vision-Language-Action Autonomous Driving}

\author{%
  \parbox{0.97\textwidth}{%
    \centering
    \mbox{Jie~Wang$^{1,2}$\thanks{Work done while interning at Xiaomi Embodied Intelligence Team.}}
    \quad \mbox{Guang~Li$^{3}$}
    \quad \mbox{Zhijian~Huang$^{3}$}
    \quad \mbox{Jinlong~Li$^{3}$}
    \quad \mbox{Chenxu~Dang$^{3}$}
    \quad \mbox{Hangjun~Ye$^{3}$}
    \quad \mbox{Yahong~Han$^{1}$\thanks{Corresponding author.}}
    \quad \mbox{Long~Chen$^{3}$}
    \\[5pt]
    \normalfont
    $^{1}$School of Artificial Intelligence, Tianjin University \\
    $^{2}$School of Artificial Intelligence, Shanxi University
    \hspace{1pt}
    $^{3}$Xiaomi EV \\[4pt]
    {\small
      \texttt{\{wangjiexy, yahong\}@tju.edu.cn},
      \hspace{1pt}
      \texttt{lguang@live.cn},
      \hspace{1pt}
      \texttt{zhijianh11@gmail.com},
      \hspace{1pt}
      \texttt{jinlong.szu@gmail.com},
      \hspace{1pt}
      \texttt{\{dangchenxu, yehangjun, chenlong37\}@xiaomi.com}
    }\\[4pt]
    {\small
      \textbf{Project Page:}
      \url{https://wj-cv.github.io/LECDrive/}
    }
  }%
}

\begin{document}

\maketitle

\vspace{-20pt}
\begin{abstract}
Autonomous driving fundamentally demands dense understanding of visual, semantic, and geometric information, yet existing Vision-Language-Action (VLA) models are typically trained with sparse supervision such as language instructions or trajectory signals. This direct mapping from dense visual observations to sparse representations forces the model to compress or even discard critical scene information essential for safe driving. We argue that this limitation does not stem from insufficient model capacity, but from the lack of explicit mechanisms that encourage the formation of dense world representations in the latent space. In this spirit, we propose a novel paradigm for driving intelligence—\textbf{Think Densely, Act Sparsely}—and instantiate it with an efficient and interpretable framework, \textbf{LECDrive} (\textbf{L}atent \textbf{E}xpert Cognitive \textbf{C}hains for Vision-Language-Action autonomous \textbf{Driv}ing). The core idea is to mimic the hierarchical perception process of humans by embedding a progressive chain of latent expert cognition within the VLA model. Concretely, we inject a compact set of task-specific latent tokens as carriers, enabling the model to internalize dense knowledge from foundational vision models, including visual representations (DINOv3), semantic structures (SAM3), and spatial geometry (DepthAnything3), while maintaining efficient inference. This interdependent cognitive chain follows a curriculum learning principle, where supervision is progressively structured from representation to semantics and geometry. Extensive experiments demonstrate that LECDrive consistently improves the performance of base VLMs across multiple autonomous driving benchmarks, covering key tasks such as object perception, state prediction, and trajectory planning. These findings suggest that constructing dense world models through structured latent expert chains is a promising direction for VLA-based autonomous driving.
\end{abstract}

\vspace{-10pt}
\section{Introduction}\label{sec1}
\vspace{-3pt}

Autonomous driving is a decision-making task that critically depends on dense world understanding \cite{chitta2022transfuser, min2024driveworld, wang2024driving}, involving fine-grained visual appearance, comprehensive semantic structure, and precise spatial geometry \cite{dang2026sparseoccvla, zhou2025hermes}. These components are not independent but are hierarchically abstracted and integrated during human perception, making autonomous driving fundamentally a dense world modeling problem \cite{shi2025drivex, li2025end}. Recent advances in large-scale Vision-Language Models (VLMs) \cite{zhu2025internvl3, bai2025qwen2, wang2024qwen2} have introduced new opportunities for addressing long-tail scenario understanding and complex semantic reasoning in autonomous driving \cite{jiang2024senna, jiang2025alphadrive, marcu2024lingoqa}. Building on this, unified Vision-Language-Action (VLA) frameworks \cite{renz2025simlingo, li2025recogdrive, luo2025adathinkdrive} adopt end-to-end learning to map multimodal inputs and instructions directly to driving trajectories in an autoregressive manner \cite{renz2025simlingo, hwang2024emma, ma2024dolphins}. However, the reliance on task-specific sparse supervision, such as language instructions or trajectory waypoints, results in a significant compression of dense visual information during training. This limitation hinders the model’s ability to fully capture complex scene structures and ultimately constrains both scene understanding and trajectory planning performance \cite{renz2025simlingo, zhou2025autovla, renz2024carllava, dang2026drivefine}.

\begin{figure}[t]
	\centering
	\includegraphics[width=1.0\columnwidth]{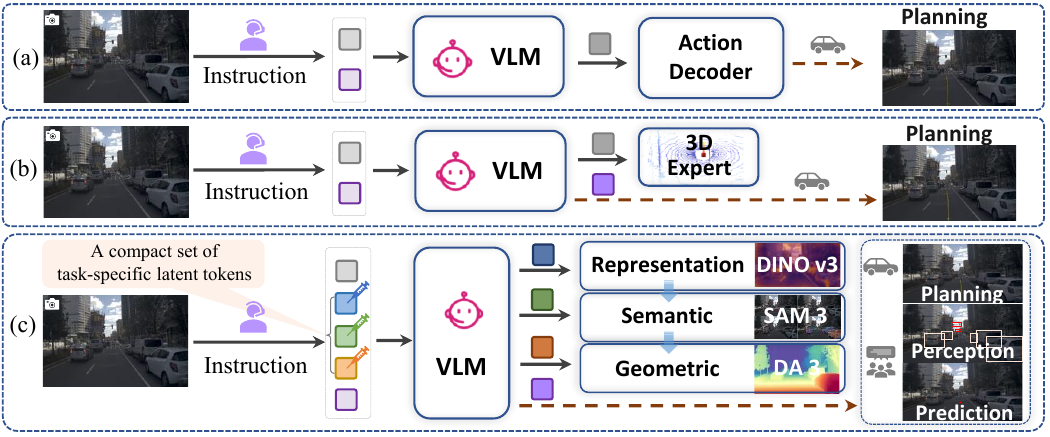}
	\DeclareGraphicsExtensions.
	\begin{center}
		\vspace{-10pt}
		\caption{Mainstream paradigms vs. LECDrive. (a) Decoupled modeling \cite{li2025recogdrive, luo2025adathinkdrive, fu2025orion}. (b) Geometry-enhanced modeling \cite{liu2025drivepi, wang2026vggdrive, luo2026last}. (c) Our Think Densely, Act Sparsely paradigm with latent expert chains, enabling progressive dense world modeling via a compact set of task-specific latent tokens.
        }\label{f1}
	\end{center}
	\vspace{-35pt}
\end{figure}

Consequently, some approaches resort to a decoupled modeling paradigm (Figure \ref{f1}(a)) by introducing an additional action decoder for trajectory prediction \cite{li2025recogdrive, luo2025adathinkdrive, fu2025orion, wang2025omnidrive}. Although this design improves planning accuracy, it decouples scene understanding from decision making, limiting the transfer of the VLM’s perception and reasoning capabilities to downstream planning. In parallel, to compensate for the inherent limitations of VLMs in spatial geometric reasoning, other methods adopt explicit geometry-enhanced pipelines (Figure \ref{f1}(b)). These approaches \cite{dang2026sparseoccvla, liu2025drivepi, wang2026vggdrive, luo2026last} incorporate 3D expert features or introduce auxiliary tasks such as point cloud or occupancy prediction on top of the visual hidden states. Although such strategies improve geometric awareness, they suffer from two key limitations. First, explicit integration of 3D features typically incurs substantial inference overhead \cite{wang2026vggdrive}. Second, naively stacking multiple geometry-related objectives often leads to task interference, hindering representation learning and resulting in suboptimal solutions \cite{liu2025drivepi, luo2026last}.  More critically, these methods \cite{liu2025drivepi, wang2026vggdrive, luo2026last} primarily focus on geometric augmentation and fail to address the fundamental issue of dense visual information loss in VLA models. In essence, autonomous driving requires not only enhanced geometric reasoning but also rich visual representations and coherent semantic structures to support robust decision making. However, this requirement is systematically overlooked in existing approaches \cite{li2025recogdrive, luo2025adathinkdrive, fu2025orion, wang2025omnidrive, liu2025drivepi, wang2026vggdrive, luo2026last}, where dense pixel-level information is prematurely compressed into sparse action representations. We argue that this limitation does not stem from insufficient model capacity, but from the absence of explicit mechanisms that enforce, or at least encourage, the construction of structured dense world representations in the latent space.

In this spirit, we propose a new modeling paradigm for autonomous driving—\textbf{Think Densely, Act Sparsely}. The central hypothesis is that if a VLA model is guided to explicitly construct dense world representations in its latent space, progressing through a cognitive chain from visual representation to semantic understanding and finally to spatial geometry, it can achieve more accurate and robust driving decisions while maintaining efficient sparse action outputs. To instantiate this idea, we introduce \textbf{LECDrive} (\textbf{L}atent \textbf{E}xpert Cognitive \textbf{C}hains for Vision-Language-Action \textbf{Driv}ing), an efficient and interpretable latent expert chain framework (Figure \ref{f1}(c)). LECDrive injects a compact set of task-specific latent tokens into the VLM, forming a progressive latent expert chain with three hierarchical stages. This design enables the model to gradually activate and internalize dense knowledge from foundational vision models, significantly enhancing its representational capacity with minimal impact on inference cost. The proposed chain follows a curriculum learning principle that mirrors the hierarchical nature of human perception. First, DINOv3 \cite{simeoni2025dinov3} provides feature-level supervision to establish robust visual representations. Second, SAM3 \cite{carion2025sam} introduces semantic segmentation signals to guide the learning of traffic-relevant semantic structures. Finally, DepthAnything3 \cite{lin2025depth} supplies geometric supervision to enhance spatial geometric understanding. These stages do not constitute independent multi-task objectives; instead, they form an interdependent cognitive chain in the latent space, where earlier stages provide increasingly rich context for subsequent stages, while later-stage supervision further refines earlier representations. \textbf{From a cognitive perspective, LECDrive establishes a latent cognitive trilogy within the VLM. Before generating driving actions, the model first learns to perceive the world like a visual feature extractor, understand it like a semantic segmenter, and finally measure it like a depth sensor.}

\begin{figure*}[!t]
  \centering
  \includegraphics[width=0.98\columnwidth]{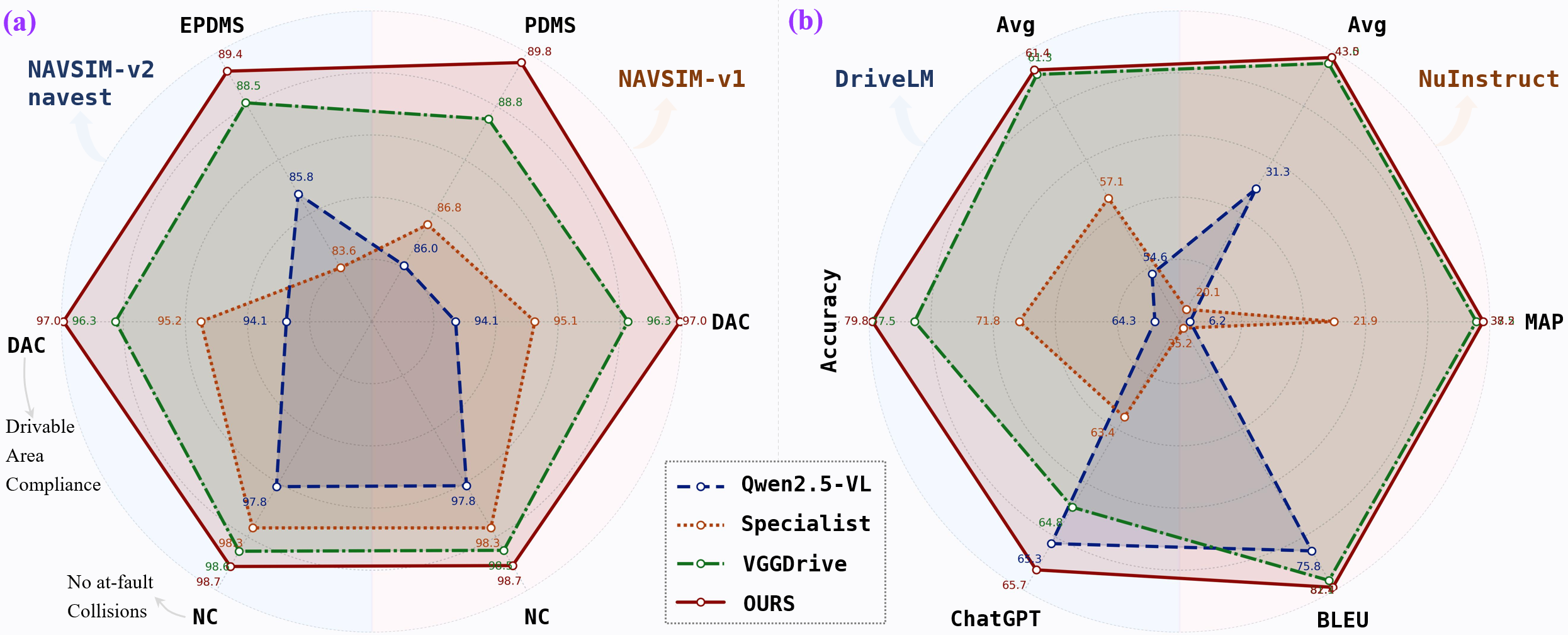}
  \vspace{-8pt}
  \caption{Quantitative comparison of LECDrive with state-of-the-art methods across three autonomous driving benchmarks, covering closed-loop planning, perception and prediction tasks.
  }\label{f2}
  \vspace{-15pt}
\end{figure*}

For trajectory prediction, we further design a hybrid formulation that combines discrete motion primitive retrieval with continuous offset refinement while preserving dense world understanding, thereby improving both the robustness and precision of trajectory planning. Extensive experiments demonstrate that LECDrive consistently and significantly improves baseline VLM performance across multiple autonomous driving benchmarks, including open-loop and closed-loop trajectory planning (Figure \ref{f2}(a)), as well as language-driven perception and prediction tasks (Figure \ref{f2}(b)). We summarize the contribution of our work as follows:

• We propose a new driving paradigm, Think Densely, Act Sparsely, which constructs dense world understanding in the latent space before generating sparse actions, providing a principled solution to dense information loss in VLA frameworks.

• We introduce an efficient and interpretable Latent Expert Cognitive Chain that injects a compact set of task-specific tokens to progressively internalize visual, semantic, and geometric knowledge, enabling dense world modeling in the latent space.

• Extensive experiments demonstrate consistent and significant performance gains of LECDrive across multiple autonomous driving benchmarks, including open-loop and closed-loop trajectory planning and language-driven perception and prediction tasks.

\vspace{-5pt}
\section{Related Works}\label{sec2}
\vspace{-3pt}
\subsection{Decoupled Action Modeling in VLA Driving}
\vspace{-2pt}


Vision-Language Models (VLMs) \cite{zhu2025internvl3, bai2025qwen2, wang2024qwen2}, with their rich world knowledge and strong reasoning capabilities, have significantly improved the generalization and interpretability of autonomous driving systems \cite{jiang2024senna, jiang2025alphadrive, marcu2024lingoqa}. From early use as high-level semantic understanding modules to end-to-end Vision-Language-Action (VLA) frameworks that autoregressively generate driving trajectories \cite{renz2025simlingo, li2025recogdrive, luo2025adathinkdrive}, the loss of dense visual information caused by sparse supervision remains a persistent bottleneck. This limitation has motivated some approaches to introduce dedicated action decoders for trajectory prediction \cite{li2025recogdrive, luo2025adathinkdrive, fu2025orion, wang2025omnidrive}. ORION \cite{fu2025orion} addresses trajectory prediction by introducing a VAE-based generative action decoder to align the reasoning and action spaces. ReCogDrive \cite{li2025recogdrive} further incorporates a diffusion-based planner, augmented with VLM-driven driving priors, as the action decoder to mitigate the mismatch between language and action spaces, enabling the generation of smooth, stable, and safe trajectories. While improving trajectory prediction accuracy, these approaches decouple scene understanding and decision making into loosely coupled stages, limiting the transfer of VLM reasoning capabilities to trajectory planning.

\vspace{-5pt}
\subsection{Explicit Geometry Augmentation for VLA Driving}
\vspace{-2pt}


To compensate for the limitations of VLMs in spatial geometric modeling, another line of work \cite{dang2026sparseoccvla, liu2025drivepi, wang2026vggdrive, luo2026last} introduces explicit geometry augmentation, leveraging 3D information or auxiliary geometric supervision to enhance the understanding of scene structure. These approaches typically incorporate multimodal signals such as point clouds, occupancy grids, or depth maps to complement or constrain visual features, thereby improving spatial perception.  VGGDrive \cite{wang2026vggdrive} enhances cross-view geometric reasoning in VLMs by injecting features from 3D expert models \cite{wang2025vggt} into the VLM architecture, thereby improving performance on multiple autonomous driving evaluation tasks. DrivePI \cite{liu2025drivepi} augments the visual representations of VLMs with point cloud and occupancy flow prediction to strengthen geometric perception, leading to improved open-loop performance on the nuScenes benchmark. However, such methods face two key challenges. First, explicitly integrating 3D information often incurs additional computational overhead, reducing inference efficiency \cite{wang2026vggdrive}. Second, naively combining multiple geometry-related objectives can lead to task interference, making it difficult to learn a unified and stable representation space \cite{liu2025drivepi, luo2026last}. More importantly, these approaches primarily focus on geometric enhancement and fail to address the fundamental issue of dense visual information loss when mapping from dense perception to sparse action representations.

\begin{figure*}[!t]
  \centering
  \includegraphics[width=1.0\columnwidth]{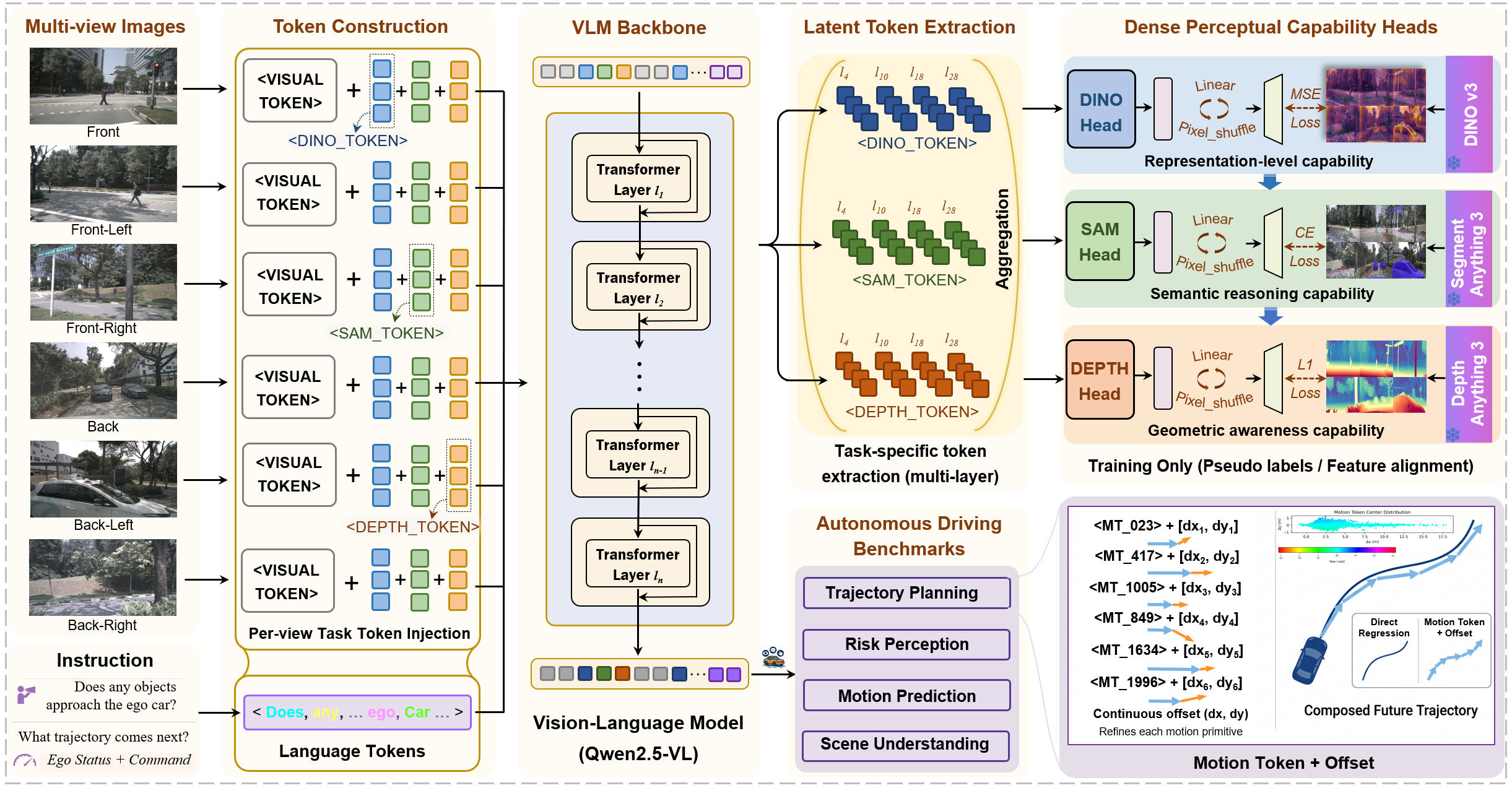}
  \vspace{-15pt}
  \caption{\textbf{Overview of LECDrive}. Three frozen dense expert models first generate task-specific pseudo-labels. A compact set of task-specific latent tokens is then injected into the visual tokens of each view. Task-specific latent embeddings from four VLM layers are aggregated and fed into dedicated perception heads for dense world prediction. For trajectory planning, LECDrive combines discrete motion primitives with continuous offset refinement to improve robustness and precision.
    }\label{f3}
    \vspace{-15pt}
\end{figure*}

\vspace{-3pt}
\section{Methodology}\label{sec3}
\vspace{-3pt}
To enable efficient dense world modeling in autonomous driving, we propose LECDrive, a latent expert cognitive chain framework that progressively internalizes visual, semantic, and geometric knowledge through a compact set of task-specific latent tokens with minimal inference overhead. As illustrated in Figure \ref{f3}, LECDrive consists of three components: 1) the VLM backbone (Qwen2.5-VL), 2) the Latent Expert Cognitive Chain, and 3) Dense Perception Heads. Details are provided below.


\vspace{-5pt}
\subsection{VLM Backbone}
\vspace{-2pt}
We adopt the Qwen2.5-VL-7B model \cite{bai2025qwen2} as the visual-language backbone for LECDrive. Qwen2.5-VL is a series of powerful multimodal large language models, it not only demonstrates exceptional visual understanding capabilities but also benefits from its open-source nature, which facilitates fine-tuning for task-specific adaptations. In our framework, the VLM receives multi-view driving images $I = \{ {I_c}\} _{c = 1}^C$ together with task-specific language instructions $L$ as input. Specifically, for benchmarks built upon the nuScenes dataset, including \textit{NuInstruct} \cite{ding2024holistic}, \textit{DriveLM} \cite{sima2024drivelm}, \textit{OmniDrive} \cite{wang2025omnidrive}, and \textit{nuScenes-Plan} \cite{caesar2020nuscenes}, we adopt six surround-view images from the current frame ($C=6$). For trajectory planning tasks on NAVSIM \cite{dauner2024navsim}, we instead use three front-view perspectives, namely front-left, front, and front-right ($C=3$). In addition, for trajectory planning benchmarks such as \textit{nuScenes-Plan} and NAVSIM, ego-vehicle states and high-level navigation commands are further incorporated into the language instructions. During instruction construction, multiple groups of compact task-specific latent tokens are inserted after the visual tokens corresponding to each view. The resulting tokens are subsequently fed into the base LLM to autoregressively generate the corresponding textual outputs. The model is optimized using the standard cross-entropy loss:
\begin{equation}
\begin{array}{c}
\mathcal{L}_{\text{CE}} =  - \sum\limits_{t = 1}^T {\log } \ {p_\theta }({y_t}|{y_{ < t}}, \ \{ {I_c}\} _{c = 1}^C, \ L),
\end{array}
\end{equation}
where ${y_t}$ is the t-th output token, and ${p_\theta }$ is the probability predicted by the model given all previous tokens and the multimodal context (i.e., images and instructions).

\vspace{-3pt}
\subsection{Latent Expert Cognitive Chain}
\vspace{-2pt}

To enable VLMs to progressively internalize dense world knowledge with minimal additional inference cost, we propose the Latent Expert Cognitive Chain in LECDrive. The core idea is to inject a compact set of task-specific latent tokens into the VLM, forming a progressive cognitive chain that sequentially internalizes visual representation (DINOv3 \cite{simeoni2025dinov3}), semantic structure (SAM3 \cite{carion2025sam}), and spatial geometry knowledge (DepthAnything3 \cite{lin2025depth} ) from foundational vision experts. Inspired by the hierarchical nature of human perception and curriculum learning, these expert stages are not optimized as independent objectives, but organized as an interdependent latent reasoning process, where higher-level supervision progressively refines lower-level representations. 

Specifically, we first inject three groups of compact task-specific latent tokens after the visual tokens of each view, each containing $n$ tokens dedicated to dense world modeling. These include \textit{<DINO\_TOKEN>} ($F_t$) for internalizing visual representation knowledge, \textit{<SAM\_TOKEN>} ($S_t$) for learning traffic-relevant semantic structures, and \textit{<DEPTH\_TOKEN>} ($D_t$) for capturing spatial geometric information. After token embedding, the corresponding latent embeddings ($\{F_e, S_e, D_e\} \in {\mathbb{R}^{B \times C \cdot n \times {D}}}$) are fed into the stacked Transformer layers of the VLM, producing hierarchical hidden states $\{ {H_i}\} _{i = 1}^{28}$. Subsequently, we explicitly extract task-specific embeddings from four Transformer layers ranging from shallow to deep and aggregate them across layers. This design is motivated by the hierarchical abstraction property naturally formed within VLM Transformers, where representations progressively evolve from local textures to global semantics. Shallow layers preserve rich fine-grained appearance details, intermediate layers capture part-level semantics and object structures, deeper layers become increasingly sensitive to object categories and spatial relationships, while the deepest layers encode high-level cross-modal reasoning capabilities. Therefore, relying on embeddings from a single layer would inevitably lead to systematic information loss under the extremely limited token budget. To obtain task-specific representations, latent embeddings ($\{F_i, S_i, D_i\} \in {\mathbb{R}^{B \times C \cdot n \times {D}}}$) are extracted from the hidden states $\{ {H_i}\} _{i = 1}^{28}$ using fixed positional ID masks corresponding to different expert tokens. Taking depth embeddings as an example.
\begin{equation}
D_i =
\begin{cases}
H_i, & M_{id}^{depth} = 1, \\
0, & \text{otherwise},
\end{cases}
\quad
i \in \{4, 10, 18, 28\},
\end{equation}
where $M_{id}^{depth}$ denotes a binary mask specific to the depth task, whose values are set to 1 at positions corresponding to the \textit{<DEPTH\_TOKEN>} and 0 otherwise. Subsequently, task-specific embeddings extracted from four Transformer layers are aggregated across layers ($\{F, S, D\} \in {\mathbb{R}^{B \times C \cdot n \times {4\cdot D}}}$) and passed to the corresponding dense world modeling heads for dense prediction.

\vspace{-3pt}
\subsection{Dense Perception Capability Heads}
\vspace{-2pt}

\begin{table*}[!t]
	\centering
	\normalsize
	\tabcolsep=0.28cm
	\renewcommand{\arraystretch}{1.0}
	\caption{Performance comparison on the NAVSIM-v1 benchmark \cite{dauner2024navsim} under closed-loop trajectory evaluation, covering both E2E methods (gray) and supervised fine-tuned VLA models (green).} \label{tab-1}
	\scalebox{0.91}{
    \begin{tabular}{lccccccccc}
        \toprule
        \multicolumn{1}{c}{\textbf{NAVSIM-v1}} & Input & NC↑   & DAC↑  & EP↑   & TTC↑  & Comf.↑ & PDMS↑ \\
        \midrule
        \rowcolor[rgb]{ .949,  .949,  .949} \multicolumn{1}{l}{PARA-Drive \cite{weng2024drive}} & \multicolumn{1}{c}{Camera} & \multicolumn{1}{c}{97.9} & \multicolumn{1}{c}{92.4} & \multicolumn{1}{c}{79.3} & \multicolumn{1}{c}{93.0} & \multicolumn{1}{c}{99.8} & \multicolumn{1}{c}{84.0} \\
        \rowcolor[rgb]{ .949,  .949,  .949} \multicolumn{1}{l}{DRAMA \cite{yuan2024drama}} & \multicolumn{1}{c}{Camera \& Lidar} & \multicolumn{1}{c}{98.19} & \multicolumn{1}{c}{95.18} & \multicolumn{1}{c}{81.33} & \multicolumn{1}{c}{94.17} & \multicolumn{1}{c}{\textbf{100.0}} & \multicolumn{1}{c}{86.87} \\
        \rowcolor[rgb]{ .949,  .949,  .949} \multicolumn{1}{l}{Hydra-MDP \cite{li2024hydra}} & \multicolumn{1}{c}{Camera \& Lidar} & \multicolumn{1}{c}{98.3} & \multicolumn{1}{c}{96.0} & \multicolumn{1}{c}{78.7} & \multicolumn{1}{c}{94.6} & \multicolumn{1}{c}{\textbf{100.0}} & \multicolumn{1}{c}{86.5} \\
        \rowcolor[rgb]{ .949,  .949,  .949} \multicolumn{1}{l}{DiffusionDrive \cite{liao2025diffusiondrive}} & \multicolumn{1}{c}{Camera \& Lidar} & \multicolumn{1}{c}{98.2} & \multicolumn{1}{c}{96.2} & \multicolumn{1}{c}{82.2} & \multicolumn{1}{c}{94.7} & \multicolumn{1}{c}{\textbf{100.0}} & \multicolumn{1}{c}{88.1} \\
        \rowcolor[rgb]{ .949,  .949,  .949} \multicolumn{1}{l}{WoTE \cite{li2025end}} & \multicolumn{1}{c}{Camera \& Lidar} & \multicolumn{1}{c}{98.5} & \multicolumn{1}{c}{96.8} & \multicolumn{1}{c}{81.9} & \multicolumn{1}{c}{94.9} & \multicolumn{1}{c}{99.9} & \multicolumn{1}{c}{88.3} \\
        \rowcolor[rgb]{ .949,  .949,  .949} \multicolumn{1}{l}{Hydra-NeXt \cite{li2024hydra}} & \multicolumn{1}{c}{Camera \& Lidar} & \multicolumn{1}{c}{98.1} & \multicolumn{1}{c}{97.7} & \multicolumn{1}{c}{81.8} & \multicolumn{1}{c}{94.6} & \multicolumn{1}{c}{\textbf{100.0}} & \multicolumn{1}{c}{88.6} \\
        \midrule
        \rowcolor[rgb]{ .886,  .937,  .855} \multicolumn{1}{l}{Baseline \cite{bai2025qwen2}} & \multicolumn{1}{c}{Camera} & \multicolumn{1}{c}{97.83} & \multicolumn{1}{c}{94.08} & \multicolumn{1}{c}{81.0} & \multicolumn{1}{c}{94.04} & \multicolumn{1}{c}{99.98} & \multicolumn{1}{c}{86.04} \\
        \rowcolor[rgb]{ .886,  .937,  .855} \multicolumn{1}{l}{ImagiDrive \cite{li2025imagidrive}} & \multicolumn{1}{c}{Camera} & \multicolumn{1}{c}{97.9} & \multicolumn{1}{c}{95.5} & \multicolumn{1}{c}{80.7} & \multicolumn{1}{c}{93.1} & \multicolumn{1}{c}{99.9} & \multicolumn{1}{c}{86.4} \\
        \rowcolor[rgb]{ .886,  .937,  .855} \multicolumn{1}{l}{AutoVLA \cite{zhou2025autovla}} & \multicolumn{1}{c}{Camera} & \multicolumn{1}{c}{96.89} & \multicolumn{1}{c}{94.43} & \multicolumn{1}{c}{75.82} & \multicolumn{1}{c}{88.06} & \multicolumn{1}{c}{99.94} & \multicolumn{1}{c}{80.54} \\
        \rowcolor[rgb]{ .886,  .937,  .855} \multicolumn{1}{l}{ReCogDrive \cite{li2025recogdrive}} & \multicolumn{1}{c}{Camera} & \multicolumn{1}{c}{98.3} & \multicolumn{1}{c}{95.1} & \multicolumn{1}{c}{81.1} & \multicolumn{1}{c}{94.3} & \multicolumn{1}{c}{\textbf{100.0}} & \multicolumn{1}{c}{86.8} \\
        \rowcolor[rgb]{ .886,  .937,  .855} \multicolumn{1}{l}{AdaThinkDrive \cite{luo2025adathinkdrive}} & \multicolumn{1}{c}{Camera} & \multicolumn{1}{c}{\textbf{98.9}} & \multicolumn{1}{c}{95.3} & \multicolumn{1}{c}{80.6} & \multicolumn{1}{c}{96.0} & \multicolumn{1}{c}{\textbf{100.0}} & \multicolumn{1}{c}{87.5} \\
        \rowcolor[rgb]{ .886,  .937,  .855} \multicolumn{1}{l}{DriveVLA-W0 \cite{li2025drivevla}} & \multicolumn{1}{c}{Camera} & \multicolumn{1}{c}{98.7} & \multicolumn{1}{c}{96.2} & \multicolumn{1}{c}{82.2} & \multicolumn{1}{c}{95.5} & \multicolumn{1}{c}{\textbf{100.0}} & \multicolumn{1}{c}{88.4} \\
        \rowcolor[rgb]{ .886,  .937,  .855} \multicolumn{1}{l}{LatentVLA \cite{xie2026latentvla}} & \multicolumn{1}{c}{Camera \& Lidar} & \multicolumn{1}{c}{98.2} & \multicolumn{1}{c}{95.9} & \multicolumn{1}{c}{79.4} & \multicolumn{1}{c}{94.8} & \multicolumn{1}{c}{99.9} & \multicolumn{1}{c}{86.6} \\
        \rowcolor[rgb]{ .886,  .937,  .855} \multicolumn{1}{l}{OneDrive \cite{zhang2026onedrive}} & \multicolumn{1}{c}{Camera} & \multicolumn{1}{c}{98.4} & \multicolumn{1}{c}{95.2} & \multicolumn{1}{c}{81.1} & \multicolumn{1}{c}{94.9} & \multicolumn{1}{c}{\textbf{100.0}} & \multicolumn{1}{c}{86.8} \\
        \rowcolor[rgb]{ .886,  .937,  .855} \multicolumn{1}{l}{LaST-VLA \cite{luo2026last}} & \multicolumn{1}{c}{Camera} & \multicolumn{1}{c}{98.7} & \multicolumn{1}{c}{95.4} & \multicolumn{1}{c}{80.5} & \multicolumn{1}{c}{95.6} & \multicolumn{1}{c}{\textbf{100.0}} & \multicolumn{1}{c}{86.8} \\
        \rowcolor[rgb]{ .886,  .937,  .855} \multicolumn{1}{l}{VGGDrive \cite{wang2026vggdrive}} & \multicolumn{1}{c}{Camera} & \multicolumn{1}{c}{98.55} & \multicolumn{1}{c}{96.3} & \multicolumn{1}{c}{82.92} & \multicolumn{1}{c}{95.59} & \multicolumn{1}{c}{99.98} & \multicolumn{1}{c}{88.76} \\
        \midrule
        \rowcolor[rgb]{ .886,  .937,  .855} \textbf{OURS} & Camera & 98.72 & \textbf{96.97} & \textbf{83.68} & \textbf{96.39} & 99.99 & \textbf{89.81} \\
        \bottomrule
		\end{tabular}%
	}
	\vspace{-0.12in}
\end{table*}%

\begin{table*}[!t]
	\centering
	\normalsize
	\tabcolsep=0.185cm
	\renewcommand{\arraystretch}{1.0}
	\caption{The performance comparison on the NAVSIM-v2 \textit{navtest} \cite{dauner2024navsim} benchmark considers both SOTA end-to-end approaches and existing VLA models under supervised fine-tuning.} \label{tab-2}
	\scalebox{0.9}{
        \begin{tabular}{lcccccccccc}
        \toprule
        \multicolumn{1}{c}{\textbf{NAVSIM-v2}} & \multicolumn{1}{c}{NC↑} & \multicolumn{1}{c}{DAC↑} & \multicolumn{1}{c}{DCC↑} & \multicolumn{1}{c}{TLC↑} & \multicolumn{1}{c}{EP↑} & \multicolumn{1}{c}{TTC↑} & \multicolumn{1}{c}{LK.↑} & \multicolumn{1}{c}{HC↑} & \multicolumn{1}{c}{EC↑} & \multicolumn{1}{c}{EPDMS↑} \\
        \midrule
        Ego-MLP \cite{li2024ego} & 93.1  & 77.9  & 92.7  & 99.6  & 86.0  & 91.5  & 89.4  & 98.3  & 85.4  & 64.0 \\
        TransFuser & 96.9  & 89.9  & 97.8  & 99.7  & 87.1  & 95.4  & 92.7  & 98.3  & 87.2  & 76.7 \\
        Hydra-MDP++ & 97.2  & \textbf{97.5} & 99.4  & 99.6  & 83.1  & 96.5  & 94.4  & 98.2  & 70.9  & 81.4 \\
        DriveSuprim \cite{yao2026drivesuprim} & 97.5  & 96.5  & 99.4  & 99.6  & \textbf{88.4} & 96.6  & 95.5  & 98.3  & 77.0  & 83.1 \\
        ReCogDrive & 98.3  & 95.2  & \textbf{99.5} & 99.8  & 87.1  & 97.5  & 96.6  & 98.3  & 86.5  & 83.6 \\
        DiffusionDrive & 98.2  & 96.3  & 99.4  & 99.8  & 87.4  & 97.4  & 97.0  & 98.3  & \textbf{87.7} & 88.3 \\
        VGGDrive & 98.56 & 96.30 & 99.38 & 99.85 & 87.85 & 98.18 & 97.29 & 98.31 & 83.40 & 88.50 \\
        \midrule
        OURS  & \textbf{98.73} & 96.97 & \textbf{99.49} & \textbf{99.88} & 87.89 & \textbf{98.43} & \textbf{97.34} & \textbf{98.34} & 84.50 & \textbf{89.44} \\
        \bottomrule
		\end{tabular}%
	}
	\vspace{-0.12in}
\end{table*}%

To enable the injected task-specific latent tokens to fully internalize dense world knowledge, we design three lightweight dense perception heads for corresponding dense prediction tasks. Meanwhile, we leverage three state-of-the-art vision expert models to generate pseudo-labels, covering visual representation features $F^{gt}$, traffic-aware semantic segmentation $S^{gt}$, and metric depth map $D^{gt}$. 

\textbf{Representation Perception Head.}
We design a lightweight representation head to align visual embeddings $\ F \in {\mathbb{R}^{B \times C \cdot n \times {4\cdot D}}}$ with the DINO feature $F^{gt}$. The input embeddings are first reshaped into a compact 2D feature map $\ F^{2d} \in {\mathbb{R}^{B \cdot C \times h \times w \times {4\cdot D}}}$, where $n = h \times w$ and $2h=w$, followed by a two-stage upsampling pipeline. In each stage, Pixel Shuffle is employed for spatial upsampling, after which the spatial tokens are linearly projected to enhance local representation capacity. 
\begin{equation} \label{eq3}
\begin{array}{c}
{X'} = \phi (PS(F^{2d})),
\end{array}
\end{equation}
where $PS(\cdot)$ denotes the Pixel Shuffle operation for spatial upsampling and $\phi (\cdot)$ denotes token flattening followed by a linear projection. The resulting high-resolution features $F^{pre}$ are finally projected into the DINO feature space $F^{gt}$ for representation alignment. To encourage the hidden representations to preserve semantic consistency with DINO features, we minimize the mean squared error (MSE) between the predicted features and the target DINO representations:
\begin{equation} \label{q4}
\mathcal{L}_{\text{DINO}}
=
\frac{1}{N}
\left\|
F^{pre} - F^{gt}
\right\|_2^2.
\end{equation}

\textbf{Semantic Perception Head.}
We design a lightweight semantic perception head to map compact task-specific embeddings $\ S \in {\mathbb{R}^{B \times C \cdot n \times {4\cdot D}}}$ into traffic-aware semantic segmentation predictions $S^{gt}$ via stacked Eq. \ref{eq3} blocks. In this task, we focus on six semantic categories that are closely related to autonomous driving, including \textit{Vehicles}, \textit{Pedestrians}, \textit{Traffic Signs}, \textit{Traffic Lights}, \textit{Road Markings}, and \textit{Road Edges}. We optimize the semantic segmentation branch using a weighted cross-entropy loss to alleviate the class imbalance issue. Specifically, lower weight is assigned to the background category, while foreground classes share identical weights.
\begin{equation}
\mathcal{L}_{\text{SAM}}
=
-
\frac{1}{N}
\sum_{i=1}^{N}
w_{y_i}
\log p_{i,y_i},
\end{equation}
where $N$ denotes the number of pixels, $y_i$ is the ground-truth label of pixel $i$, $p_{i,y_i}$ denotes the predicted probability for the corresponding class, and $w_{y_i}$ is the class-specific weighting factor. We set the background class weight to $0.2$ and assign a weight of $1.0$ to all remaining classes.

\textbf{Geometric Perception Head.}
Since DepthAnything3 \cite{lin2025depth} supports camera-parameter-conditioned depth estimation, and such information is naturally available in autonomous driving scenarios, camera parameters are incorporated when generating depth pseudo-labels. Meanwhile, a lightweight depth estimation head is designed to map depth-specific embeddings into metric depth predictions, implemented using stacked Eq. \ref{eq3} blocks. We supervise the depth estimation branch using a masked Smooth $L_1$ loss. Only valid depth regions are involved in optimization, where depth values are constrained within a predefined range.
\begin{equation}
\mathcal{L}_{\text{DEPTH}}
=
\frac{
\sum_{i=1}^{N}
M_i \cdot
\operatorname{SmoothL1}
\left(
D_i^{pre},
D_i^{gt}
\right)
}{
\sum_{i=1}^{N} M_i + \epsilon
},
\end{equation}
where $D_i^{pre}$ and $D_i^{gt}$ denote the predicted and ground-truth depth values at pixel $i$, respectively, $M_i$ is the validity mask, and $\epsilon$ is a small constant for numerical stability. We set $M_i = 1$ when $0 < D_i^{gt} \leq 60$, and $M_i = 0$ otherwise. The overall training objective is defined as:
\begin{equation}
\mathcal{L}
= \lambda_1 \cdot \mathcal{L}_{\text{CE}} + \lambda_2 \cdot \mathcal{L}_{\text{DINO}} + \lambda_3 \cdot \mathcal{L}_{\text{SAM}} + \lambda_4 \cdot \mathcal{L}_{\text{DEPTH}},
\end{equation}
where $\{ {\lambda_i}\} _{i = 1}^4$ represents the weighting factor for each objective in the joint training process.

\begin{table*}[!t]
	\centering
	\normalsize
	\tabcolsep=0.04cm
	\renewcommand{\arraystretch}{1.0}
	\caption{Comparison with sota methods on the NuInstruct dataset \cite{ding2024holistic} for risk object perception and state prediction. * indicates $\max((\text{Accuracy} + \text{MAP} + \text{BLEU} - \text{MAE})/4, 0)$.} \label{tab-3}
	\scalebox{0.905}{
    \begin{tabular}{ccccccccc}
    \toprule
        \multicolumn{1}{c}{\multirow{2}[4]{*}{Dataset}} & \multirow{2}[4]{*}{Metrics} & \multicolumn{3}{c}{Zero-shot Models} & \multicolumn{4}{c}{Fine-tuned Models} \\
        \cmidrule(lr){3-5} \cmidrule(lr){6-9}    
        & \multicolumn{1}{c}{} & \multicolumn{1}{c}{GPT-4o} \cite{hurst2024gpt} & \multicolumn{1}{c}{RoboTron} \cite{huang2025robotron} & \multicolumn{1}{c}{Qwen2.5-VL} & \multicolumn{1}{c}{Baseline} & \multicolumn{1}{c}{InMLLM} \cite{ding2024holistic} & \multicolumn{1}{c}{VGGDrive} & \multicolumn{1}{c}{OURS} \\
        \midrule
        \multicolumn{1}{c}{\multirow{5}[2]{*}{NuInstruct}} & \textbf{MAP↑} & 0       & 0     & 0  & 6.15  & 21.93 & 37.49 & \textbf{38.25} \\
        & Accuracy↑ & 10.64  & 2.57  & 0.63 & 47.71 & 32.48 & 56.37 & \textbf{57.02} \\
        & MAE↓  & 9.93  & 19.36 & 24.10  & 4.35  & 9.08  & \textbf{3.08} & 3.57 \\
        & BLEU↑ & 7.08   & 8.06  & 5.56 & 75.75 & 35.20 & 81.13 & \textbf{82.38} \\
        & \textbf{Average*↑} & 1.95    & 0     & 0 & 31.32 & 20.13 & 42.98 & \textbf{43.52} \\
        \bottomrule
		\end{tabular}%
	}
	\vspace{-0.15in}
\end{table*}%

\begin{table*}[!t]
	\centering
	\normalsize
	\tabcolsep=0.01cm
	\renewcommand{\arraystretch}{1.0}
	\caption{The performance comparison on the DriveLM dataset \cite{sima2024drivelm} is conducted against existing state-of-the-art (SOTA) methods to evaluate risk object perception, action prediction, and planning.} \label{tab-4}
	\scalebox{0.87}{
    \begin{tabular}{cccccccccc}
    \toprule
    \multicolumn{1}{c}{\multirow{2}[4]{*}{Dataset}} & \multirow{2}[4]{*}{Metrics} & \multicolumn{8}{c}{Fine-tuned Models} \\
        \cmidrule{3-10}     & \multicolumn{1}{c}{} & \multicolumn{1}{c}{Baseline} & \multicolumn{1}{c}{DriveLM} & \multicolumn{1}{c}{TMLMM} \cite{ishaq2025tracking} & \multicolumn{1}{c}{OminiD} \cite{wang2025omnidrive} & \multicolumn{1}{c}{FSDrive} \cite{zeng2025FSDdrive} & \multicolumn{1}{c}{IVL4Drive} \cite{li2024driving}  & \multicolumn{1}{c}{VGGDrive} & \multicolumn{1}{c}{OURS} \\
        \midrule
        \multicolumn{1}{c}{\multirow{5}[2]{*}{DriveLM}} & Accuracy↑ & 64.35 & 52.30 & 59.60 & 70.00 & 71.77 & 73.39 & 77.50 & \textbf{79.81} \\
        & ChatGPT↑ & 65.30 & 55.35 & 58.44 & 65.00 & 63.42 & 65.25 & 64.76 & \textbf{65.69} \\
        & Language↑ & 44.90 & 45.13 & 46.38 & \multicolumn{1}{c}{-} & 52.77 & 48.56 & \textbf{56.69} & 56.53 \\
        & Match↑ & 34.54 & 35.73 & 35.73 & 37.00 & 39.19 & 47.65 & \textbf{49.77} & 47.73 \\
        & \textbf{Average↑} & 54.59 & 47.20 & 51.50 & 56.00 & 57.12 & 60.02 & 61.26 & \textbf{61.41} \\
    \bottomrule
		\end{tabular}%
	}
	\vspace{-0.15in}
\end{table*}%

\vspace{-5pt}
\section{Experiments}\label{sec4}
\vspace{-3pt}
\subsection{Implementation Details}\label{sec4.1}
\vspace{-3pt}

In our experiments, we adopt Qwen2.5-VL-7B \cite{bai2025qwen2} as the base Vision-Language Model. The proposed latent expert cognitive chain is instantiated with three progressive expert modules: a feature representation expert based on DINOv3 \cite{simeoni2025dinov3}, a semantic expert based on SAM 3 \cite{carion2025sam}, and a geometric expert based on DepthAnything 3 \cite{lin2025depth}. Specifically, DINOv3 employs a ViT-based model with 840M parameters pretrained on the large-scale web dataset LVD-1689M. For geometric modeling, we adopt the DepthAnything3 (DA3NESTED-GIANT-LARGE) variant with 1.4B parameters. During training, the learning rate is set to $5 \times 10^{-5}$ and the model is trained for 5 epochs. The number of task-specific latent tokens for each expert $n$ is set to 8. The loss weights for different components are set to $\lambda_1=1$, $\lambda_2=0.5$, $\lambda_3=0.5$, and $\lambda_4=1.5$. All experiments are conducted on 8 NVIDIA H200 GPUs. For trajectory planning, we cluster trajectory data into 2048 motion primitives, transforming direct waypoint prediction into a hybrid formulation of discrete motion primitive retrieval with continuous offset refinement. Details are provided in the Appendix \ref{secA3}.

\vspace{-5pt}
\subsection{Dataset and Metric}
\begin{figure*}[!t]
  \centering
  \includegraphics[width=1.0\columnwidth]{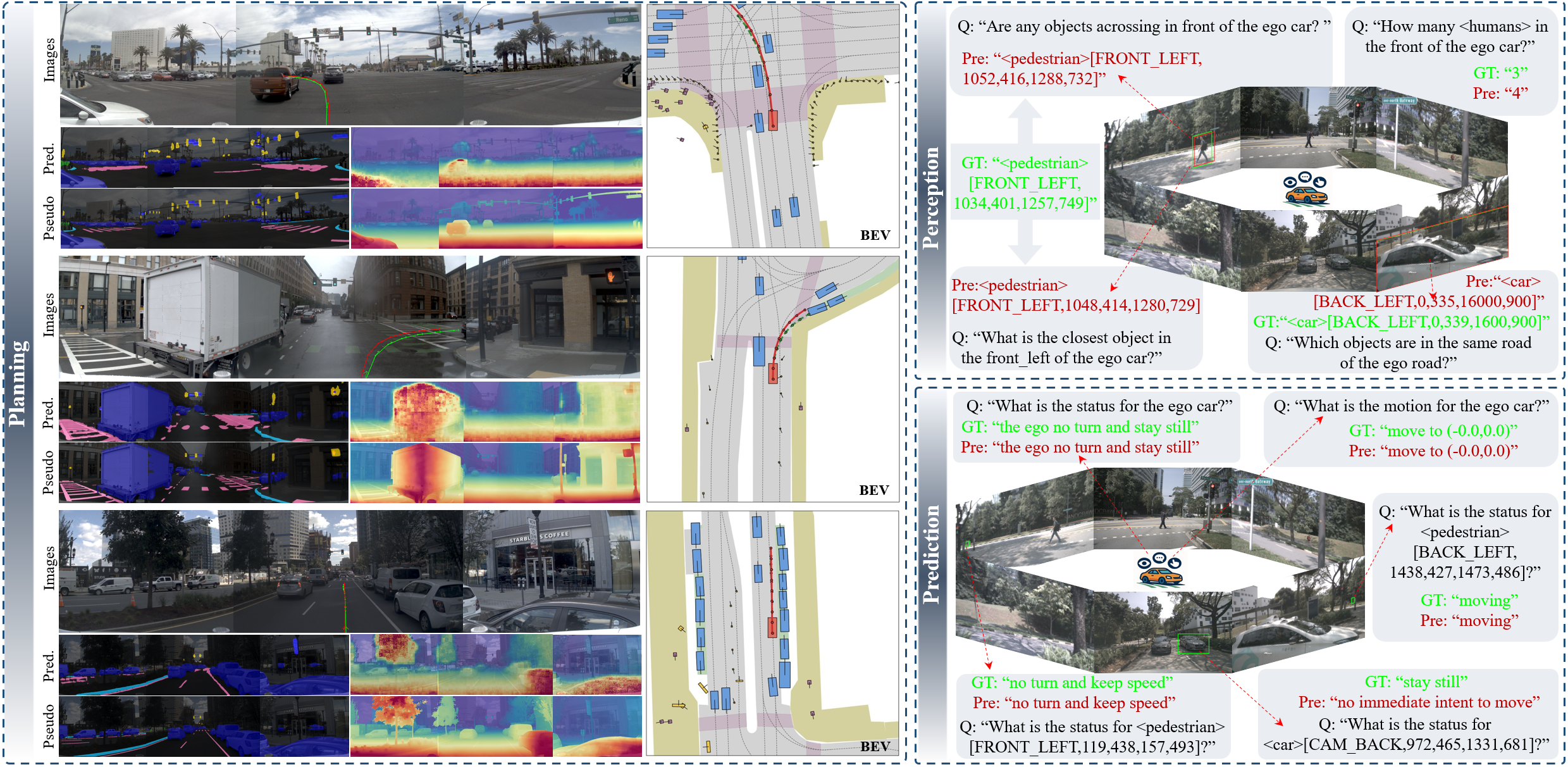}
  \vspace{-15pt}
  \caption{Visualization of LECDrive across various driving tasks. Besides trajectory planning, we showcase the model’s internal dense world modeling results, including semantic segmentation and depth estimation generated by dedicated heads, along with risk object perception and state prediction.}
    \label{f4}
     \vspace{-15pt}
\end{figure*}

\vspace{-5pt}
To validate the effectiveness of the proposed Think Densely, Act Sparsely paradigm—constructing dense world understanding before generating sparse driving actions—and the latent expert cognitive chain architecture, we conduct experiments across various autonomous driving tasks on five mainstream benchmarks. These benchmarks include language evaluation benchmarks: NuInstruct \cite{ding2024holistic}, DriveLM \cite{sima2024drivelm}, and Omnidrive \cite{wang2025omnidrive}, as well as two trajectory planning tasks: NuScenes-Plan \cite{caesar2020nuscenes} and NAVSIM \cite{dauner2024navsim}. The NuInstruct and DriveLM datasets contain numerous cases for evaluating risk object perception, state and motion prediction (e.g., Figure \ref{f4}). The Omnidrive dataset primarily focuses on scene description tasks related to captioning in driving scenarios. The NuScenes-Plan and NAVSIM datasets focus on open-loop and closed-loop trajectory planning tasks, respectively. To ensure a fair comparison, we evaluate LECDrive on five mainstream datasets using their standard metrics. Detailed dataset and metric descriptions are provided in Appendix \ref{secA1}.

\vspace{-5pt}
\subsection{Performance Comparison}
\vspace{-5pt}
\textbf{NAVSIM.} As reported in Table \ref{tab-1}, the closed-loop trajectory comparison is shown. 1) Compared to AdaThinkDrive, a representative decoupled approach that relies on an additional action decoder for trajectory prediction, LECDrive achieves a \textbf{+2.31} improvement in PDMS. 2) Compared with VGGDrive, which enhances spatial perception by injecting 3D features into VLMs, LECDrive still achieves a \textbf{+1.05} gain in PDMS. Under the more stringent NAVSIM v2 \textit{navtest} benchmark (Table \ref{tab-2}), LECDrive further outperforms VGGDrive by \textbf{+0.94} in EPDMS. See appendix \ref{secA2} for more details. 

\textbf{NuInstruct.} As shown in Table \ref{tab-3}, the base VLM performs poorly on tasks such as risk object perception (MAP) and state prediction, and exhibits only marginal gains even after fine-tuning on task-specific datasets. In contrast, the proposed LECDrive achieves substantial improvements on the key MAP and Accuracy metrics, with gains of \textbf{+32.10} and \textbf{+9.31}, respectively. 

\textbf{DriveLM.} This benchmark is crucial for evaluating a model's capability in object perception, action prediction and planning. As reported in Table \ref{tab-4}, LECDrive substantially surpasses the baseline, improving the Accuracy and Average metrics by \textbf{15.46} and \textbf{6.82}, respectively.

\textbf{OmniDrive and NuScenes-plan.} These two benchmarks primarily serve as complementary evaluations in our experiments. Detailed descriptions and comparisons are provided in Appendix \ref{secA2}.

\textbf{LECDrive’s consistent advantages across multiple benchmarks suggest that a progressive latent expert cognitive chain, which enhances dense world modeling in VLMs without incurring additional inference cost, is both effective and promising for VLA-based autonomous driving.}

\textbf{Visualization Analysis.} Figure \ref{f4} visualizes the performance of LECDrive across multiple autonomous driving tasks. For trajectory planning, we further present the model’s dense world modeling results, including semantic segmentation and depth estimation alongside expert pseudo-labels. With only 8 task-specific latent tokens, LECDrive effectively enables dense world understanding in VLMs. Moreover, the model also demonstrates strong performance on risk object perception and state prediction tasks. Details are provided in the Appendix \ref{secA2}

\vspace{-2pt}
\subsection{Ablation Study}
\vspace{-2pt}

We perform extensive ablation studies on the NAVSIM-v1 \cite{dauner2024navsim} closed-loop benchmark to systematically evaluate the effectiveness of the proposed architecture and its core components.

\textbf{Effectiveness of the Latent Expert Cognitive Chain.} Table \ref{tab-5} presents the ablation study on the effectiveness of the three chained expert modules in LECDrive for closed-loop trajectory planning. Specifically, comparing ID-3 and ID-4 shows that introducing the dense visual representation expert, DINOv3, yields a \textbf{+0.58} improvement in PDMS. Building upon this, incorporating the semantic structure expert SAM 3 further improves performance by \textbf{+0.24}. When the geometric expert DepthAnything 3 is additionally integrated, the model achieves another significant gain of \textbf{+0.50}, reaching a state-of-the-art PDMS of \textbf{89.81}. Comparing ID-6 and ID-7 indicates that visual representation and semantic experts exhibit shared characteristics when internalized by VLMs, yet removing either expert leads to performance degradation, highlighting their complementary contributions. These results demonstrate that progressively introducing dense expert modules in a chained manner—mirroring the hierarchical nature of human perception—enables VLMs to incrementally internalize dense world knowledge, leading to consistent performance improvements.

\begin{table*}[!t]
	\centering
	\normalsize
	\tabcolsep=0.14cm
	\renewcommand{\arraystretch}{1.0}
    \caption{Ablation study on the effectiveness of the proposed key components for closed-loop planning.} \label{tab-5}
    \scalebox{0.91}{
    \begin{tabular}{ccccccccccccc}
    \toprule
    \multicolumn{1}{c}{\multirow{1}[0]{*}{ID}} & Baseline & MT & Offset & DINO v3 & SAM 3 & DA 3 & \multicolumn{1}{c}{NC↑} & \multicolumn{1}{c}{DAC↑} & \multicolumn{1}{c}{EP↑} & \multicolumn{1}{c}{TTC↑} & \multicolumn{1}{c}{Comf.↑} & \multicolumn{1}{c}{PDMS↑} \\
    \cmidrule{1-13} 
    1     & $\checkmark$     &  &  &  &  &  & 97.83 & 94.08 & 81.00 & 94.04 & 99.98 & 86.04 \\
    2     & $\checkmark$     & $\checkmark$     &   &   &   &   & 98.44 & 94.95 & 81.45 & 95.72 & 99.99 & 87.44 \\
    3     & $\checkmark$     & $\checkmark$     & $\checkmark$     &   &   &   & 98.62 & 96.06 & 82.38 & 95.78 & \textbf{100.00} & 88.49 \\
    4     & $\checkmark$     & $\checkmark$     & $\checkmark$     & $\checkmark$     &   &   & 98.53 & 96.57 & 82.99 & 95.97 & 99.99 & 89.07 \\
    5     & $\checkmark$     & $\checkmark$     & $\checkmark$     & $\checkmark$     & $\checkmark$     &   & 98.68 & 96.76 & 83.27 & 95.97 & 99.98 & 89.31 \\
    6     & $\checkmark$     & $\checkmark$     & $\checkmark$     &   & $\checkmark$     & $\checkmark$     & 98.65 & 96.85 & 83.54 & 96.08 & 99.99 & 89.53 \\
    7     & $\checkmark$     & $\checkmark$     & $\checkmark$     & $\checkmark$     &   & $\checkmark$     & 98.69 & 96.80 & 83.48 & 96.29 & \textbf{100.00} & 89.57 \\
    8     & $\checkmark$     & $\checkmark$     & $\checkmark$     & $\checkmark$     & $\checkmark$     & $\checkmark$     & \textbf{98.72} & \textbf{96.97} & \textbf{83.68} & \textbf{96.39} & 99.99 & \textbf{89.81} \\
    \bottomrule
		\end{tabular}%
	}
	\vspace{-0.15in}
\end{table*}

\begin{table*}[!t]
	\centering
	\normalsize
	\tabcolsep=0.266cm
	\renewcommand{\arraystretch}{1.0}
    \caption{Ablation Study on Architectural Variants and Key Hyperparameters in LECDrive.} \label{tab-6}
    \scalebox{0.95}{
        \begin{tabular}{clcccccc}
        \toprule
        \multicolumn{1}{c}{\multirow{1}[1]{*}{ID}} & \multirow{1}[1]{*}{Methods} & \multicolumn{1}{c}{NC↑} & \multicolumn{1}{c}{DAC↑} & \multicolumn{1}{c}{EP↑} & \multicolumn{1}{c}{TTC↑} & \multicolumn{1}{c}{Comf.↑} & \multicolumn{1}{c}{PDMS↑} \\
        \cmidrule{1-8} 
        1     & Baseline & 97.83 & 94.08 & 81.00 & 94.04 & 99.98 & 86.04 \\
        2     & Visual tokens + Task heads & 98.58 & 96.28 & 82.73 & 95.82 & 99.99 & 88.73 \\
        3     & Task tokens ($l_{28}$) + Task heads  & 98.59 & 96.78 & 83.14 & 95.94 & 99.99 & 89.26 \\
        4     & Task tokens = 2 & 98.57 & 96.54 & 82.93 & 96.16 & 99.99 & 89.13 \\
        5     & Qwen2.5-VL-3B & 98.66 & 96.66 & 83.19 & 95.99 & \textbf{100.00} & 89.21 \\
        6     & OURS  & \textbf{98.72} & \textbf{96.97} & \textbf{83.68} & \textbf{96.39} & 99.99 & \textbf{89.81} \\
        \bottomrule
		\end{tabular}%
	}
	\vspace{-0.15in}
\end{table*}

\textbf{Effectiveness of the Hybrid Trajectory Formulation.} Table \ref{tab-5} also presents an ablation study on the proposed hybrid trajectory formulation. Replacing direct coordinate regression with discrete motion primitive retrieval (ID-2) improves PDMS by \textbf{+1.40}, demonstrating the benefit of structured discrete trajectory modeling, although this formulation alone remains less competitive with sota methods (Table \ref{tab-1}). Further introducing continuous offset refinement boosts PDMS to \textbf{88.49}, making the model competitive with strong expert-designed methods (88.49 \textit{vs.} 88.76 for VGGDrive \cite{wang2026vggdrive}). These results suggest that discrete primitives provide robustness, while continuous offsets recover fine-grained precision. Overall, the hybrid formulation effectively balances robustness and accuracy, significantly improving trajectory prediction performance. Additional details are provided in Appendix \ref{secA3}.


\textbf{Effectiveness of the Architectural Variants and Key Hyperparameters.} Table \ref{tab-6} presents ablation studies on architectural variants and key hyperparameters of LECDrive. Removing task-specific latent tokens (ID-2) causes a significant performance drop (\textbf{-1.08} PDMS), indicating severe task interference without explicit latent specialization. Using only final-layer latent tokens without multi-layer aggregation (ID-3) further reduces performance by \textbf{-0.55} PDMS, demonstrating the importance of integrating shallow and deep VLM features. ID-4 and ID-5 further analyze the effects of latent token quantity and base model scale, both of which influence overall performance.


\noindent\textbf{Disentangling Expert Knowledge from Trajectory Modeling.}
To isolate the contribution of latent expert knowledge internalization from the hybrid trajectory formulation, we evaluate the expert modules without MT+Offset. As shown in Table \ref{tab-10}(a), DINOv3, SAM 3, and DepthAnything 3 individually improve PDMS from 86.04 to 87.98, 88.17, and 88.47, respectively. Combining all three experts further achieves 89.15 PDMS, yielding a substantial gain of +3.11 over the baseline. These results demonstrate that dense expert knowledge internalization improves planning performance independently of the hybrid trajectory representation.

\noindent\textbf{Effect of Expert Ordering.}
We further investigate the importance of the proposed causal expert ordering. As shown in
Table \ref{tab-10}(b), changing the representation--semantics--geometry order to representation--geometry--semantics or semantics--representation--geometry decreases PDMS from 89.81 to 89.18 and 89.66, respectively. These results indicate that the ordering of expert-specific latent tokens influences knowledge integration, supporting the effectiveness of the proposed progressive cognitive hierarchy.

\noindent\textbf{Computational Efficiency.}
Without MT+Offset, LECDrive increases training time from 2.6 to 5.4 hours per epoch, primarily due to online dense pseudo-label generation using three frozen vision foundation models, rather than the lightweight perception heads. This overhead can be substantially reduced by precomputing and caching expert pseudo-labels. Importantly, all expert models and dense prediction heads are discarded during inference, leaving only compact expert-specific latent tokens that account for 2.7\% of the total input tokens. Consequently, inference time increases only marginally from 0.80 to 0.85 seconds, demonstrating that LECDrive retains internalized dense world knowledge without substantial inference overhead.

\begin{table*}[!t]
	\centering
	\normalsize
	\tabcolsep=0.266cm
	\renewcommand{\arraystretch}{1.0}
    \caption{Further ablations on expert knowledge internalization and causal ordering on NAVSIM-v1.} \label{tab-10}
    \scalebox{0.95}{
        \begin{tabular}{lccccc}
        \toprule
        Configuration & NC$\uparrow$ & DAC$\uparrow$
        & EP$\uparrow$ & TTC$\uparrow$ & PDMS$\uparrow$ \\
        \midrule
        \multicolumn{6}{l}{\textit{(a) Expert supervision without MT+Offset}} \\
        Baseline             & 97.83 & 94.08 & 81.00 & 94.04 & 86.04 \\
        + DINOv3             & 98.38 & 95.49 & 82.71 & 95.01 & 87.98 \\
        + SAM 3              & 98.21 & 95.80 & 83.03 & 97.85 & 88.17 \\
        + DepthAnything 3    & 98.60 & 96.01 & 82.39 & 95.81 & 88.47 \\
        + All Experts        & 98.59 & 96.51 & 82.97 & 96.05 & \textbf{89.15} \\
        \midrule
        \multicolumn{6}{l}{\textit{(b) Expert ordering with MT+Offset}} \\
        DINO $\rightarrow$ DEPTH $\rightarrow$ SAM
                             & 98.58 & 96.54 & 82.93 & 96.26 & 89.18 \\
        SAM $\rightarrow$ DINO $\rightarrow$ DEPTH
                             & 98.63 & 96.93 & 83.59 & 96.20 & 89.66 \\
        DINO $\rightarrow$ SAM $\rightarrow$ DEPTH
                             & \textbf{98.72} & \textbf{96.97}
                             & \textbf{83.68} & \textbf{96.39}
                             & \textbf{89.81} \\
        \bottomrule
		\end{tabular}%
	}
	\vspace{-0.15in}
\end{table*}

\vspace{-5pt}
\section{Limitations}\label{sec5}
\vspace{-2pt}
Although LECDrive effectively alleviates the mismatch between dense perception and sparse supervision under the \textbf{Think Densely, Act Sparsely} paradigm, several limitations remain. 1) The method relies on high-quality pseudo-labels and external expert models, whose capacity and generalization inherently bound performance. 2) A limited number of latent tokens may introduce representation bottlenecks (Figure \ref{f4}), while the scaling behavior of token quantity is not explored due to computational constraints. 3) The current design focuses on single-frame or weak temporal modeling, leaving dynamic consistency and unified multi-view scene understanding underexplored. Future work will explore more expressive latent representations, adaptive token allocation, and unified spatio-temporal modeling to improve robustness and generalization in complex real-world scenarios.

\vspace{-5pt}
\section{Conclusion}\label{sec6}
\vspace{-2pt}
In this paper, we revisit the fundamental limitation of VLA models for autonomous driving, namely the mismatch between dense perception and sparse supervision, which leads to the loss of critical scene information. To address this issue, we propose a new paradigm, Think Densely, Act Sparsely, which emphasizes constructing dense world representations in the latent space prior to action generation. Building upon this paradigm, we introduce LECDrive, an efficient and interpretable latent expert cognitive chain framework that progressively internalizes visual, semantic, and geometric knowledge via a compact set of task-specific tokens. Extensive experiments across multiple autonomous driving benchmarks demonstrate that LECDrive consistently improves performance on trajectory planning, perception, and prediction tasks. More broadly, our findings suggest that explicitly modeling structured dense representations in the latent space is a promising direction for advancing VLA-based autonomous driving. We hope this work inspires future research toward unified and efficient frameworks that bridge perception and action through richer internal world modeling.

\begin{ack}
This work was supported by the National Natural Science Foundation of China (NSFC) under Grant No. 62376186.

The authors declare no competing interests.
\end{ack}

\bibliographystyle{IEEEtran}
\bibliography{main}

@String(PAMI = {IEEE Trans. Pattern Anal. Mach. Intell.})

@String(CVPR= {IEEE Conf. Comput. Vis. Pattern Recog.})

@String(ICCV= {Int. Conf. Comput. Vis.})

@String(ECCV= {Eur. Conf. Comput. Vis.})

@String(NIPS= {Adv. Neural Inform. Process. Syst.})

@String(ICLR = {Int. Conf. Learn. Represent.})

@String(AAAI = {AAAI})

@String(PAMI  = {IEEE TPAMI})

@String(CVPR  = {CVPR})

@String(ICCV  = {ICCV})

@String(ECCV  = {ECCV})

@String(NIPS  = {NeurIPS})

@String(ICLR  = {ICLR})

@inproceedings{min2024driveworld,
  title={Driveworld: 4d pre-trained scene understanding via world models for autonomous driving},
  author={Min, Chen and Zhao, Dawei and Xiao, Liang and Zhao, Jian and Xu, Xinli and Zhu, Zheng and Jin, Lei and Li, Jianshu and Guo, Yulan and Xing, Junliang and others},
  booktitle=CVPR,
  pages={15522--15533},
  year={2024}
}

@inproceedings{wang2024driving,
  title={Driving into the future: Multiview visual forecasting and planning with world model for autonomous driving},
  author={Wang, Yuqi and He, Jiawei and Fan, Lue and Li, Hongxin and Chen, Yuntao and Zhang, Zhaoxiang},
  booktitle=CVPR,
  pages={14749--14759},
  year={2024}
}

@inproceedings{shi2025drivex,
  title={Drivex: Omni scene modeling for learning generalizable world knowledge in autonomous driving},
  author={Shi, Chen and Shi, Shaoshuai and Sheng, Kehua and Zhang, Bo and Jiang, Li},
  booktitle=CVPR,
  pages={28599--28609},
  year={2025}
}

@inproceedings{zhou2025hermes,
  title={Hermes: A unified self-driving world model for simultaneous 3d scene understanding and generation},
  author={Zhou, Xin and Liang, Dingkang and Tu, Sifan and Chen, Xiwu and Ding, Yikang and Zhang, Dingyuan and Tan, Feiyang and Zhao, Hengshuang and Bai, Xiang},
  booktitle=ICCV,
  pages={27817--27827},
  year={2025}
}

@article{dang2026sparseoccvla,
  title={SparseOccVLA: Bridging Occupancy and Vision-Language Models via Sparse Queries for Unified 4D Scene Understanding and Planning},
  author={Dang, Chenxu and Wang, Jie and Li, Guang and Hou, Zhiwen and You, Zihan and Ye, Hangjun and Ma, Jie and Chen, Long and Wang, Yan},
  journal={arXiv preprint arXiv:2601.06474},
  year={2026}
}

@article{zhu2025internvl3,
  title={Internvl3: Exploring advanced training and test-time recipes for open-source multimodal models},
  author={Zhu, Jinguo and Wang, Weiyun and Chen, Zhe and Liu, Zhaoyang and Ye, Shenglong and Gu, Lixin and Tian, Hao and Duan, Yuchen and Su, Weijie and Shao, Jie and others},
  journal={arXiv preprint arXiv:2504.10479},
  year={2025}
}

@article{bai2025qwen2,
  title={Qwen2. 5-vl technical report},
  author={Bai, Shuai and Chen, Keqin and Liu, Xuejing and Wang, Jialin and Ge, Wenbin and Song, Sibo and Dang, Kai and Wang, Peng and Wang, Shijie and Tang, Jun and others},
  journal={arXiv preprint arXiv:2502.13923},
  year={2025}
}

@article{wang2024qwen2,
  title={Qwen2-vl: Enhancing vision-language model's perception of the world at any resolution},
  author={Wang, Peng and Bai, Shuai and Tan, Sinan and Wang, Shijie and Fan, Zhihao and Bai, Jinze and Chen, Keqin and Liu, Xuejing and Wang, Jialin and Ge, Wenbin and others},
  journal={arXiv preprint arXiv:2409.12191},
  year={2024}
}

@article{jiang2024senna,
  title={Senna: Bridging large vision-language models and end-to-end autonomous driving},
  author={Jiang, Bo and Chen, Shaoyu and Liao, Bencheng and Zhang, Xingyu and Yin, Wei and Zhang, Qian and Huang, Chang and Liu, Wenyu and Wang, Xinggang},
  journal={arXiv preprint arXiv:2410.22313},
  year={2024}
}

@article{jiang2025alphadrive,
  title={Alphadrive: Unleashing the power of vlms in autonomous driving via reinforcement learning and reasoning},
  author={Jiang, Bo and Chen, Shaoyu and Zhang, Qian and Liu, Wenyu and Wang, Xinggang},
  journal={arXiv preprint arXiv:2503.07608},
  year={2025}
}

@inproceedings{marcu2024lingoqa,
  title={Lingoqa: Visual question answering for autonomous driving},
  author={Marcu, Ana-Maria and Chen, Long and H{\"u}nermann, Jan and Karnsund, Alice and Hanotte, Benoit and Chidananda, Prajwal and Nair, Saurabh and Badrinarayanan, Vijay and Kendall, Alex and Shotton, Jamie and others},
  booktitle={ECCV},
  pages={252--269},
  year={2024},
  organization={Springer}
}

@article{renz2024carllava,
  title={Carllava: Vision language models for camera-only closed-loop driving},
  author={Renz, Katrin and Chen, Long and Marcu, Ana-Maria and H{\"u}nermann, Jan and Hanotte, Benoit and Karnsund, Alice and Shotton, Jamie and Arani, Elahe and Sinavski, Oleg},
  journal={arXiv preprint arXiv:2406.10165},
  year={2024}
}

@inproceedings{fu2025orion,
  title={ORION: A Holistic End-to-End Autonomous Driving Framework by Vision-Language Instructed Action Generation},
  author={Haoyu Fu and Diankun Zhang and Zongchuang Zhao and Jianfeng Cui and Dingkang Liang and Chong Zhang and Dingyuan Zhang and Hongwei Xie and Bing Wang and Xiang Bai},
  booktitle=ICCV,
  year={2025}
}

@article{hwang2024emma,
  title={Emma: End-to-end multimodal model for autonomous driving},
  author={Hwang, Jyh-Jing and Xu, Runsheng and Lin, Hubert and Hung, Wei-Chih and Ji, Jingwei and Choi, Kristy and Huang, Di and He, Tong and Covington, Paul and Sapp, Benjamin and others},
  journal={arXiv preprint arXiv:2410.23262},
  year={2024}
}

@inproceedings{renz2025simlingo,
  title={Simlingo: Vision-only closed-loop autonomous driving with language-action alignment},
  author={Renz, Katrin and Chen, Long and Arani, Elahe and Sinavski, Oleg},
  booktitle=CVPR,
  pages={11993--12003},
  year={2025}
}

@inproceedings{ma2024dolphins,
  title={Dolphins: Multimodal language model for driving},
  author={Ma, Yingzi and Cao, Yulong and Sun, Jiachen and Pavone, Marco and Xiao, Chaowei},
  booktitle=ECCV,
  pages={403--420},
  year={2024},
  organization={Springer}
}

@article{liu2025drivepi,
  title={DrivePI: Spatial-aware 4D MLLM for Unified Autonomous Driving Understanding, Perception, Prediction and Planning},
  author={Liu, Zhe and Huang, Runhui and Yang, Rui and Yan, Siming and Wang, Zining and Hou, Lu and Lin, Di and Bai, Xiang and Zhao, Hengshuang},
  journal={arXiv preprint arXiv:2512.12799},
  year={2025}
}

@article{simeoni2025dinov3,
  title={Dinov3},
  author={Sim{\'e}oni, Oriane and Vo, Huy V and Seitzer, Maximilian and Baldassarre, Federico and Oquab, Maxime and Jose, Cijo and Khalidov, Vasil and Szafraniec, Marc and Yi, Seungeun and Ramamonjisoa, Micha{\"e}l and others},
  journal={arXiv preprint arXiv:2508.10104},
  year={2025}
}

@article{carion2025sam,
  title={Sam 3: Segment anything with concepts},
  author={Carion, Nicolas and Gustafson, Laura and Hu, Yuan-Ting and Debnath, Shoubhik and Hu, Ronghang and Suris, Didac and Ryali, Chaitanya and Alwala, Kalyan Vasudev and Khedr, Haitham and Huang, Andrew and others},
  journal={arXiv preprint arXiv:2511.16719},
  year={2025}
}

@article{lin2025depth,
  title={Depth anything 3: Recovering the visual space from any views},
  author={Lin, Haotong and Chen, Sili and Liew, Junhao and Chen, Donny Y and Li, Zhenyu and Shi, Guang and Feng, Jiashi and Kang, Bingyi},
  journal={arXiv preprint arXiv:2511.10647},
  year={2025}
}

@inproceedings{wang2025vggt,
  title={Vggt: Visual geometry grounded transformer},
  author={Wang, Jianyuan and Chen, Minghao and Karaev, Nikita and Vedaldi, Andrea and Rupprecht, Christian and Novotny, David},
  booktitle={Proceedings of the Computer Vision and Pattern Recognition Conference},
  pages={5294--5306},
  year={2025}
}

@inproceedings{caesar2020nuscenes,
  title={nuscenes: A multimodal dataset for autonomous driving},
  author={Caesar, Holger and Bankiti, Varun and Lang, Alex H and Vora, Sourabh and Liong, Venice Erin and Xu, Qiang and Krishnan, Anush and Pan, Yu and Baldan, Giancarlo and Beijbom, Oscar},
  booktitle=CVPR,
  pages={11621--11631},
  year={2020}
}

@article{dauner2024navsim,
  title={Navsim: Data-driven non-reactive autonomous vehicle simulation and benchmarking},
  author={Dauner, Daniel and Hallgarten, Marcel and Li, Tianyu and Weng, Xinshuo and Huang, Zhiyu and Yang, Zetong and Li, Hongyang and Gilitschenski, Igor and Ivanovic, Boris and Pavone, Marco and others},
  journal=NIPS,
  volume={37},
  pages={28706--28719},
  year={2024}
}

@article{chitta2022transfuser,
  title={Transfuser: Imitation with transformer-based sensor fusion for autonomous driving},
  author={Chitta, Kashyap and Prakash, Aditya and Jaeger, Bernhard and Yu, Zehao and Renz, Katrin and Geiger, Andreas},
  journal=PAMI,
  volume={45},
  number={11},
  pages={12878--12895},
  year={2022},
  publisher={IEEE}
}

@inproceedings{weng2024drive,
  title={Para-drive: Parallelized architecture for real-time autonomous driving},
  author={Weng, Xinshuo and Ivanovic, Boris and Wang, Yan and Wang, Yue and Pavone, Marco},
  booktitle=CVPR,
  pages={15449--15458},
  year={2024}
}

@article{yuan2024drama,
  title={Drama: An efficient end-to-end motion planner for autonomous driving with mamba},
  author={Yuan, Chengran and Zhang, Zhanqi and Sun, Jiawei and Sun, Shuo and Huang, Zefan and Lee, Christina Dao Wen and Li, Dongen and Han, Yuhang and Wong, Anthony and Tee, Keng Peng and others},
  journal={arXiv preprint arXiv:2408.03601},
  year={2024}
}

@article{li2024hydra,
  title={Hydra-mdp: End-to-end multimodal planning with multi-target hydra-distillation},
  author={Li, Zhenxin and Li, Kailin and Wang, Shihao and Lan, Shiyi and Yu, Zhiding and Ji, Yishen and Li, Zhiqi and Zhu, Ziyue and Kautz, Jan and Wu, Zuxuan and others},
  journal={arXiv preprint arXiv:2406.06978},
  year={2024}
}

@inproceedings{liao2025diffusiondrive,
  title={Diffusiondrive: Truncated diffusion model for end-to-end autonomous driving},
  author={Liao, Bencheng and Chen, Shaoyu and Yin, Haoran and Jiang, Bo and Wang, Cheng and Yan, Sixu and Zhang, Xinbang and Li, Xiangyu and Zhang, Ying and Zhang, Qian and others},
  booktitle=CVPR,
  pages={12037--12047},
  year={2025}
}

@inproceedings{li2025end,
  title={End-to-end driving with online trajectory evaluation via bev world model},
  author={Li, Yingyan and Wang, Yuqi and Liu, Yang and He, Jiawei and Fan, Lue and Zhang, Zhaoxiang},
  booktitle={Proceedings of the IEEE/CVF International Conference on Computer Vision},
  pages={27137--27146},
  year={2025}
}

@article{li2025imagidrive,
  title={ImagiDrive: A Unified Imagination-and-Planning Framework for Autonomous Driving},
  author={Li, Jingyu and Zhang, Bozhou and Jin, Xin and Deng, Jiankang and Zhu, Xiatian and Zhang, Li},
  journal={arXiv preprint arXiv:2508.11428},
  year={2025}
}

@inproceedings{zhou2025autovla,
  title={AutoVLA: A Vision-Language-Action Model for End-to-End Autonomous Driving with Adaptive Reasoning and Reinforcement Fine-Tuning},
  author={Zhou, Zewei and Cai, Tianhui and Zhao, Seth Z and Zhang, Yun and Huang, Zhiyu and Zhou, Bolei and Ma, Jiaqi},
  booktitle=NIPS,
  year={2025}
}

@inproceedings{li2025recogdrive,
  title={Recogdrive: A reinforced cognitive framework for end-to-end autonomous driving},
  author={Li, Yongkang and Xiong, Kaixin and Guo, Xiangyu and Li, Fang and Yan, Sixu and Xu, Gangwei and Zhou, Lijun and Chen, Long and Sun, Haiyang and Wang, Bing and others},
  booktitle=ICLR,
  year={2026}
}

@article{luo2025adathinkdrive,
  title={AdaThinkDrive: Adaptive Thinking via Reinforcement Learning for Autonomous Driving},
  author={Luo, Yuechen and Li, Fang and Xu, Shaoqing and Lai, Zhiyi and Yang, Lei and Chen, Qimao and Luo, Ziang and Xie, Zixun and Jiang, Shengyin and Liu, Jiaxin and others},
  journal={arXiv preprint arXiv:2509.13769},
  year={2025}
}

@inproceedings{li2025drivevla,
  title={DriveVLA-W0: World models amplify data scaling law in autonomous driving},
  author={Li, Yingyan and Shang, Shuyao and Liu, Weisong and Zhan, Bing and Wang, Haochen and Wang, Yuqi and Chen, Yuntao and Wang, Xiaoman and An, Yasong and Tang, Chufeng and others},
  journal={arXiv preprint arXiv:2510.12796},
  booktitle=ICLR,
  year={2026}
}

@inproceedings{wang2026vggdrive,
  title={Vggdrive: Empowering vision-language models with cross-view geometric grounding for autonomous driving},
  author={Wang, Jie and Li, Guang and Huang, Zhijian and Dang, Chenxu and Ye, Hangjun and Han, Yahong and Chen, Long},
  booktitle=CVPR,
  year={2026}
}

@article{luo2026last,
  title={Last-vla: Thinking in latent spatio-temporal space for vision-language-action in autonomous driving},
  author={Luo, Yuechen and Li, Fang and Xu, Shaoqing and Ji, Yang and Zhang, Zehan and Wang, Bing and Shen, Yuannan and Cui, Jianwei and Chen, Long and Chen, Guang and others},
  journal={arXiv preprint arXiv:2603.01928},
  year={2026}
}

@article{zhang2026onedrive,
  title={OneDrive: Unified Multi-Paradigm Driving with Vision-Language-Action Models},
  author={Zhang, Yiwei and Chen, Xuesong and Gao, Jin and Wang, Hanshi and Ge, Fudong and Hu, Weiming and Shi, Shaoshuai and Zhang, Zhipeng},
  journal={arXiv preprint arXiv:2604.17915},
  year={2026}
}

@article{xie2026latentvla,
  title={LatentVLA: Efficient Vision-Language Models for Autonomous Driving via Latent Action Prediction},
  author={Xie, Chengen and Sun, Bin and Li, Tianyu and Wu, Junjie and Hao, Zhihui and Lang, XianPeng and Li, Hongyang},
  journal={arXiv preprint arXiv:2601.05611},
  year={2026}
}

@inproceedings{li2024ego,
  title={Is ego status all you need for open-loop end-to-end autonomous driving?},
  author={Li, Zhiqi and Yu, Zhiding and Lan, Shiyi and Li, Jiahan and Kautz, Jan and Lu, Tong and Alvarez, Jose M},
  booktitle={Proceedings of the IEEE/CVF Conference on Computer Vision and Pattern Recognition},
  pages={14864--14873},
  year={2024}
}

@inproceedings{yao2026drivesuprim,
  title={Drivesuprim: Towards precise trajectory selection for end-to-end planning},
  author={Yao, Wenhao and Li, Zhenxin and Lan, Shiyi and Wang, Zi and Sun, Xinglong and Alvarez, Jose M and Wu, Zuxuan},
  booktitle={Proceedings of the AAAI Conference on Artificial Intelligence},
  volume={40},
  number={14},
  pages={11910--11918},
  year={2026}
}

@inproceedings{ding2024holistic,
  title={Holistic autonomous driving understanding by bird's-eye-view injected multi-modal large models},
  author={Ding, Xinpeng and Han, Jianhua and Xu, Hang and Liang, Xiaodan and Zhang, Wei and Li, Xiaomeng},
  booktitle=CVPR,
  pages={13668--13677},
  year={2024}
}

@article{hurst2024gpt,
  title={Gpt-4o system card},
  author={Hurst, Aaron and Lerer, Adam and Goucher, Adam P and Perelman, Adam and Ramesh, Aditya and Clark, Aidan and Ostrow, AJ and Welihinda, Akila and Hayes, Alan and Radford, Alec and others},
  journal={arXiv preprint arXiv:2410.21276},
  year={2024}
}

@inproceedings{huang2025robotron,
  title={RoboTron-Drive: All-in-One Large Multimodal Model for Autonomous Driving},
  author={Huang, Zhijian and Feng, Chengjian and Yan, Feng and Xiao, Baihui and Jie, Zequn and Zhong, Yujie and Liang, Xiaodan and Ma, Lin},
  booktitle=ICCV,
  pages={8011--8021},
  year={2025}
}

@article{ishaq2025tracking,
  title={Tracking meets large multimodal models for driving scenario understanding},
  author={Ishaq, Ayesha and Lahoud, Jean and Khan, Fahad Shahbaz and Khan, Salman and Cholakkal, Hisham and Anwer, Rao Muhammad},
  journal={arXiv preprint arXiv:2503.14498},
  year={2025}
}

@inproceedings{zeng2025FSDdrive,
  title={FutureSightDrive: Thinking Visually with Spatio-Temporal CoT for Autonomous Driving},
  author={Zeng, Shuang and Chang, Xinyuan and Xie, Mengwei and Liu, Xinran and Bai, Yifan and Pan, Zheng and Xu, Mu and Wei, Xing},
  booktitle=NIPS,
  year={2025}
}

@article{li2024driving,
  title={Driving with internvl: Oustanding champion in the track on driving with language of the autonomous grand challenge at cvpr 2024},
  author={Li, Jiahan and Li, Zhiqi and Lu, Tong},
  journal={arXiv preprint arXiv:2412.07247},
  year={2024}
}

@inproceedings{sima2024drivelm,
  title={Drivelm: Driving with graph visual question answering},
  author={Sima, Chonghao and Renz, Katrin and Chitta, Kashyap and Chen, Li and Zhang, Hanxue and Xie, Chengen and Bei{\ss}wenger, Jens and Luo, Ping and Geiger, Andreas and Li, Hongyang},
  booktitle=ECCV,
  pages={256--274},
  year={2024},
  organization={Springer}
}

@inproceedings{wang2025omnidrive,
  title={Omnidrive: A holistic vision-language dataset for autonomous driving with counterfactual reasoning},
  author={Wang, Shihao and Yu, Zhiding and Jiang, Xiaohui and Lan, Shiyi and Shi, Min and Chang, Nadine and Kautz, Jan and Li, Ying and Alvarez, Jose M},
  booktitle=CVPR,
  pages={22442--22452},
  year={2025}
}

@inproceedings{hu2023planning,
  title={Planning-oriented autonomous driving},
  author={Hu, Yihan and Yang, Jiazhi and Chen, Li and Li, Keyu and Sima, Chonghao and Zhu, Xizhou and Chai, Siqi and Du, Senyao and Lin, Tianwei and Wang, Wenhai and others},
  booktitle=CVPR,
  pages={17853--17862},
  year={2023}
}

@inproceedings{jiang2023vad,
  title={Vad: Vectorized scene representation for efficient autonomous driving},
  author={Jiang, Bo and Chen, Shaoyu and Xu, Qing and Liao, Bencheng and Chen, Jiajie and Zhou, Helong and Zhang, Qian and Liu, Wenyu and Huang, Chang and Wang, Xinggang},
  booktitle=ICCV,
  pages={8340--8350},
  year={2023}
}

@inproceedings{dang2026drivefine,
  title={Drivefine: Refining-augmented masked diffusion vla for precise and robust driving},
  author={Dang, Chenxu and Ang, Sining and Li, Yongkang and Tian, Haochen and Wang, Jie and Li, Guang and Ye, Hangjun and Ma, Jie and Chen, Long and Wang, Yan},
  booktitle={ECCV},
  year={2026},
  organization={Springer}
}
\newpage

\appendix
\section{Appendix}\label{secA}
To further complement the main content of this paper, we provide additional materials to elaborate on several key aspects of our method.Specifically, Sec. \ref{secA1} presents a more detailed description of the application of LECDrive across five mainstream autonomous driving benchmarks, along with their corresponding evaluation metrics. Sec. \ref{secA2} provides additional quantitative comparisons between LECDrive and representative methods on the nuScenes open-loop trajectory planning benchmark and the OmniDrive language evaluation benchmark. Sec. \ref{secA3} further introduces the hybrid formulation for trajectory planning, which combines discrete motion primitive retrieval with continuous offset refinement, along with supplementary visualizations.

\subsection{Dataset and Metric}\label{secA1}
To assess the performance of LECDrive across various attributes within the autonomous driving domain, we conduct evaluations on five prominent benchmarks. These benchmarks cover task-specific scenarios, including scene understanding, risk object perception, action and state prediction, and trajectory planning.

\textbf{NAVSIM-v1.} This dataset \cite{dauner2024navsim} is a real-world, planning-focused dataset derived from OpenScene, which is a compact version of nuPlan, the largest publicly available annotated driving dataset (1,192 and 136 scenarios for training and testing). NAVSIM is designed to evaluate the performance of autonomous driving systems in complex, dynamic scenarios. It utilizes a combination of eight cameras providing a 360° field of view (FOV) and a merged LiDAR point cloud from five sensors.  NAVSIM specifically targets challenging driving situations with dynamic driving intentions, while deliberately excluding simpler, static scenarios such as stationary scenes or constant-speed driving. The NAVSIM benchmark provides a nonreactive simulation environment and employs the Predictive Driver Model Score (PDMS) as its closed-loop planning metric:
\begin{equation}
\begin{array}{c}
PDMS = NC \times DAC \times \left( {\frac{{5 \times EP + 5 \times TTC + 2 \times C}}{{12}}} \right)
\end{array}
\end{equation}
where PDMS integrates five sub-metrics: No At-Fault Collision (NC), Drivable Area Compliance (DAC), Time-to Collision (TTC), Comfort (C), and Ego Progress (EP) to produce a comprehensive closed-loop planning score. 

\textbf{NAVSIM-v2.} In NAVSIM-v2 navtest, the Enhanced Planning and Driving Metric System (EPDMS) extends the original PDMS by introducing several new evaluation components, including two weighted metrics—Lane Keeping (LK) and Extended Comfort (EC)—two multiplicative metrics—Driving Direction Compliance (DDC) and Traffic Light Compliance (TLC)—as well as a false-positive penalty filtering mechanism. The Lane Keeping (LK) subscore penalizes prolonged deviations from the lane centerline. In addition, NAVSIM v2 places greater emphasis on driving comfort by extending the original Comfort (C) metric to History Comfort (HC), which evaluates the consistency between the planned trajectory and the vehicle’s motion history. The newly introduced Extended Comfort (EC) further measures temporal consistency by comparing trajectory outputs and corresponding dynamic states across consecutive frames, where discrepancies in acceleration, jerk, and related dynamics lead to lower scores. Moreover, the Driving Direction Compliance (DDC) and Traffic Light Compliance (TLC) metrics expand the range of detectable and penalized driving violations. To reduce false-positive penalties, NAVSIM v2 disables penalties in cases where human drivers would also violate traffic rules. This design prevents unfair penalization in scenarios where rule violations are necessary to achieve valid driving objectives, such as temporarily entering an oncoming lane to bypass obstacles.
\begin{equation}
\begin{array}{c}
EPDMS = NC \times DAC \times DDC \times TLC \times \left( {\frac{{5 \times EP + 5 \times TTC + 2 \times LK + 2 \times HC + 2 \times EC}}{{16}}} \right)
\end{array}
\end{equation}

\textbf{NuInstruct.} This dataset \cite{ding2024holistic} samples a total of 11,850 keyframes from 850 videos within the NuScenes dataset. It includes a wide range of challenging samples, such as cross-view risk perception, target distance estimation, agent and ego state prediction, target motion prediction, and reasoning tasks. The dataset effectively reflects the model's ability to perform cross-view understanding and analyze 3D geometric scene information. NuInstruct uses a variety of metrics to evaluate different tasks, including Mean Absolute Error (MAE) for regression tasks, accuracy for classification tasks, Mean Average Precision (mAP) for detection tasks, and a rule-based BLEU metric for captioning tasks.

\textbf{DriveLM.} This dataset \cite{sima2024drivelm} is built upon the real-world driving dataset nuScenes and covers interactive scenarios involving vehicles, pedestrians, and traffic infrastructure on urban roads. The keyframes focus on moments that mark changes in driving intentions, such as acceleration, deceleration, and turning. The data samples not only include ``what objects are currently present" (cross-view perception), but also encompass ``how objects will move in the future" (action $\&$ states prediction), ``what the vehicle should do" (planning), ``specific behavior classifications" (e.g., fast driving straight, slow right turn), and ``trajectory coordinates" (motion). This dataset similarly reflects the model's ability to perform cross-view perception and analyze 3D geometric scene understanding. DriveLM implements four evaluation metrics: accuracy, LLM score, language rule-based evaluation, and match score.

\textbf{OmniDrive.} This dataset \cite{wang2025omnidrive} is also built upon the nuScenes dataset and primarily focuses on scene description, attention-based question answering, counterfactual reasoning question answering, decision planning question answering, and general dialogue. Due to the limitations in the availability of evaluation data and metrics, we selected scene description and general dialogue from this dataset to train and evaluate the model's ability to understand autonomous driving scene captions. The dataset primarily includes three language evaluation metrics: BLEU, CIDEr, and ROUGE, with the ``average" representing the mean of the three metrics.

\textbf{nuScenes.} We conduct open-loop trajectory planning experiments on the challenging public nuScenes dataset \cite{li2024ego}, which contains 1,000 driving scenes, each lasting approximately 20 seconds. The scene images are captured by six cameras, providing a 360° horizontal field of view (FOV), with keyframes annotated at a frequency of 2Hz. We follow the standard training and testing setup, adhering to the evaluation metrics used in OmniDrive. Among these metrics \cite{wang2025omnidrive, li2024ego}, L2 and Collision Rate are widely used, and we additionally introduce the Intersection Rate, which calculates the rate of collisions or intersections between the predicted trajectories and curbs (road boundaries).

\textbf{Data Usage Details.} For both training and inference, we use multi-view images from the current frame. Specifically, 6-view surrounding images are used for benchmarks built on the nuScenes dataset, including NuInstruct \cite{ding2024holistic}, DriveLM \cite{sima2024drivelm}, OmniDrive \cite{wang2025omnidrive}, and nuScenes-Plan. For NAVSIM, a closed-loop trajectory planning benchmark, we adopt 3-view front-facing images. In trajectory planning tasks on nuScenes-Plan \cite{caesar2020nuscenes} and NAVSIM \cite{dauner2024navsim}, ego-vehicle states (e.g., velocity and acceleration) and high-level commands (e.g., Go Straight, Turn Left, Turn Right) are incorporated into the prompts.

\begin{table*}[!t]
	\centering
	\normalsize
	\tabcolsep=0.17cm
	\renewcommand{\arraystretch}{1.0}
	\caption{Performance comparison on nuScenes open-loop planning, with metrics from BEV-Planner's reproduced results \cite{li2024ego}.} \label{tab-7}
	\scalebox{0.91}{
    \begin{tabular}{lccccccccccccc}
    \toprule
    \multicolumn{2}{c}{\multirow{2}[4]{*}{\textbf{nuScenes}}} & \multicolumn{4}{c}{L2 ↓} & \multicolumn{4}{c}{Collision (\%) ↓} & \multicolumn{4}{c}{Intersection (\%) ↓} \\
    \cmidrule(lr){3-6} \cmidrule(lr){7-10} \cmidrule(lr){11-14} 
    \multicolumn{2}{c}{} & \multicolumn{1}{c}{1s} & \multicolumn{1}{c}{2s} & \multicolumn{1}{c}{3s} & \multicolumn{1}{c}{avg} & \multicolumn{1}{c}{1s} & \multicolumn{1}{c}{2s} & \multicolumn{1}{c}{3s} & \multicolumn{1}{c}{avg} & \multicolumn{1}{c}{1s} & \multicolumn{1}{c}{2s} & \multicolumn{1}{c}{3s} & \multicolumn{1}{c}{avg} \\
    \midrule
    UniAD \cite{hu2023planning} & \multicolumn{1}{c}{\multirow{3}[2]{*}{E2E}} & 0.20  & 0.42  & 0.75  & 0.46  & 0.02  & 0.25  & 0.84  & 0.37  & 0.20  & 1.33  & 3.24  & 1.59 \\
    VAD-Base \cite{jiang2023vad} &       & 0.17  & 0.34  & 0.60  & 0.37  & 0.04  & 0.27  & 0.67  & 0.33  & 0.21  & 2.13  & 5.06  & 2.47 \\
    BEV-Planner &       & 0.16  & 0.32  & 0.57  & 0.35  & 0.00  & 0.29  & 0.73  & 0.34  & 0.35  & 2.62  & 6.51  & 3.16 \\
    \midrule
    Baseline & \multicolumn{1}{c}{\multirow{5}[2]{*}{VLA}} & \textbf{0.14} & 0.34  & 0.60  & 0.36  & 0.06  & 0.21  & 0.96  & 0.41  & 0.66  & 2.58  & 5.43  & 2.89 \\
    OmniDrive-Q &       & \textbf{0.14} & 0.29  & 0.55  & 0.33  & \textbf{0.00} & 0.13  & 0.78  & 0.30  & 0.56  & 2.48  & 5.96  & 3.00 \\
    OmniDrive-L &       & 0.15  & 0.36  & 0.70  & 0.40  & 0.06  & 0.27  & 0.72  & 0.35  & \textbf{0.49} & \textbf{1.99} & 4.86  & 2.45 \\
    VGGDrive &       & \textbf{0.14} & \textbf{0.28} & \textbf{0.51} & \textbf{0.31} & 0.02  & \textbf{0.10} & 0.55  & 0.22  & 0.63  & 2.27  & \textbf{4.02} & 2.31 \\
    LECDrive  &       & \textbf{0.14} & 0.30  & 0.53  & 0.32  & 0.02  & 0.12  & \textbf{0.50} & \textbf{0.21} & 0.58  & 2.04  & 4.17  & \textbf{2.26} \\
    \bottomrule
	\end{tabular}%
	}
\end{table*}%

\begin{table*}[!t]
	\centering
	\normalsize
	\tabcolsep=0.13cm
	\renewcommand{\arraystretch}{1.0}
    \caption{The performance comparison on the OmniDrive dataset \cite{wang2025omnidrive} focuses on caption-related tasks in autonomous driving scenarios, where base VLMs excel.} \label{tab-8}
    \scalebox{0.91}{
    \begin{tabular}{ccccccccc}
    \toprule
        \multicolumn{1}{c}{\multirow{2}[4]{*}{Dataset}} & \multirow{2}[4]{*}{Metrics} & \multicolumn{2}{c}{Zero-shot Models} & \multicolumn{5}{c}{Fine-tuned Models} \\
        \cmidrule(lr){3-4} \cmidrule(lr){5-9}   
        & \multicolumn{1}{c}{} & \multicolumn{1}{c}{GPT-4o} & \multicolumn{1}{c}{RoboTron} & \multicolumn{1}{c}{Ominidrive} & Hermes \cite{zhou2025hermes} & \multicolumn{1}{c}{Baseline} & \multicolumn{1}{c}{VGGDrive} & \multicolumn{1}{c}{LECDrive} \\
        \midrule
        \multicolumn{1}{r}{\multirow{4}[2]{*}{Omnidrive}} & BLEU↑ & 10.91 & 20.30 & 38.00 & -     & 37.29 & 37.58 & \textbf{37.73} \\
        & CIDEr↑ & 24.42 & 34.33 & 68.60 & \multicolumn{1}{c}{74.10} & 86.29 & \textbf{86.57} & 86.34 \\
        & ROUGE↑ & 22.34  & 23.67 & 32.60 & \multicolumn{1}{c}{32.70} & 34.33 & 34.40 & \textbf{34.42} \\
        & \textbf{Average↑} & 19.22 & 26.10 & 46.40 & -  & 52.64 & \textbf{52.85} & 52.83 \\
        \bottomrule
		\end{tabular}%
	}
\end{table*}%

\subsection{Performance Comparison}
\label{secA2}

\textbf{NAVSIM.} As reported in Table \ref{tab-1}, the closed-loop trajectory comparison is shown. 1) Compared to AdaThinkDrive, a representative decoupled approach that relies on an additional action decoder for trajectory prediction, LECDrive achieves a \textbf{+2.31} improvement in PDMS. 2) Compared with the representative method VGGDrive, which enhances spatial perception by injecting 3D features into VLMs, LECDrive still achieves a \textbf{+1.05} gain in PDMS. On the more stringent NAVSIM v2 navtest benchmark (Table \ref{tab-2}), LECDrive still achieves a \textbf{+0.94} absolute improvement in EPDMS over the representative method VGGDrive. Without exaggeration, LECDrive achieves state-of-the-art performance for VLM-based autoregressive trajectory generation, with a PDMS of 89.81 and an EPDMS of 89.44. These findings highlight that enhancing dense world modeling in VLMs via a progressive latent expert cognitive chain offers a more effective and efficient solution for trajectory prediction than relying solely on auxiliary action decoders.

\textbf{NuInstruct.} Compared to the baseline model (Qwen2.5-VL-7B) fully fine-tuned on the specific dataset (Table \ref{tab-3}), the proposed LECDrive achieves substantial improvements on the key MAP and Accuracy metrics, with gains of \textbf{+32.10} and \textbf{+9.31}, respectively. Language evaluation metrics, such as BLEU, also show consistent improvements, with an average gain of \textbf{+12.2} across four metrics. Compared to the representative method VGGDrive, LECDrive further improves MAP and Accuracy metric by \textbf{+0.76} and \textbf{+0.65}, respectively.

\textbf{NuScenes-plan.} Table \ref{tab-7} presents a quantitative comparison between LECDrive and recent state-of-the-art methods on the nuScenes open-loop trajectory planning benchmark. As shown in the table, LECDrive achieves highly competitive performance compared to the latest approaches. Notably, it demonstrates clear advantages in collision-related metrics and interactions with road boundaries. These improvements can be attributed to our chain-structured dense world modeling design, which enables the model to better perceive both static and dynamic agents.

\textbf{OmniDrive.} Table \ref{tab-8} reports the quantitative results on the OmniDrive language evaluation benchmark. Prior work \cite{wang2025omnidrive} has shown that transferring VLMs to autonomous driving often introduces a trade-off between language capability and trajectory planning performance. While the base VLM already achieves strong results after fine-tuning on this dataset, our results demonstrate that LECDrive significantly improves driving-related capabilities—including planning, perception, and prediction—without compromising its language performance on captioning tasks.

\textbf{Visualization Analysis.}  Figure \ref{f4} visualizes the performance of LECDrive across multiple autonomous driving tasks. For trajectory planning, we additionally present the model’s internal dense world modeling results, including semantic segmentation and depth estimation generated by dedicated heads, together with the corresponding pseudo-labels provided by expert models. These visualizations show that, with only 8 task-specific latent tokens, LECDrive is able to effectively activate the VLM’s capability to understand and model dense world representations. Although a gap remains between the model predictions and expert pseudo-labels, our objective is not to strictly reconstruct the ground-truth dense world, but to enable the VLM to internalize structured dense representations that benefit downstream driving tasks. Moreover, LECDrive also demonstrates satisfactory performance on risk object perception and state prediction tasks.

\textbf{Effectiveness of the Hybrid Trajectory Formulation.} Table \ref{tab-5} also presents an ablation study on the proposed hybrid trajectory formulation, which combines discrete motion primitive retrieval with continuous offset refinement. Compared to the baseline (ID-1), replacing direct coordinate regression with discrete motion primitive retrieval (ID-2) yields a \textbf{+1.40} improvement in PDMS, demonstrating the benefit of modeling trajectories in a structured discrete space. However, as shown in Table \ref{tab-1}, this formulation alone is not yet competitive with state-of-the-art methods. Adding continuous offset refinement further boosts performance to \textbf{88.49} PDMS, making the model competitive with strong expert-designed methods (88.49 \textit{vs.} 88.76 for VGGDrive \cite{wang2026vggdrive}). These results suggest that discrete primitives provide robustness, while continuous offsets recover fine-grained precision. Overall, the hybrid formulation effectively balances robustness and accuracy, significantly improving trajectory prediction performance. 

\textbf{Effectiveness of the Architectural Variants and Key Hyperparameters.} Table \ref{tab-6} presents ablation studies on architectural variants and key hyperparameters of LECDrive. ID-2 removes the task-specific latent tokens used for internalizing multi-expert representations and instead directly relies on the VLM’s visual features. Under this setting, significant performance degradation is observed (\textbf{-1.08} PDMS), indicating that the absence of task-specific latent tokens leads to severe task interference and hinders effective knowledge specialization. ID-3 uses only the final-layer latent tokens without performing multi-layer aggregation of task-specific tokens within the VLM. This configuration results in a performance drop of \textbf{-0.55} PDMS, suggesting that explicitly integrating features from earlier layers of the VLM is beneficial for capturing fine-grained representations and improving overall performance. ID-4 and ID-5 further examine the impact of the number of latent tokens and the scale of the base model, showing that both factors affect performance to varying degrees.

\begin{figure*}[!t]
  \centering
  \includegraphics[width=1.0\columnwidth]{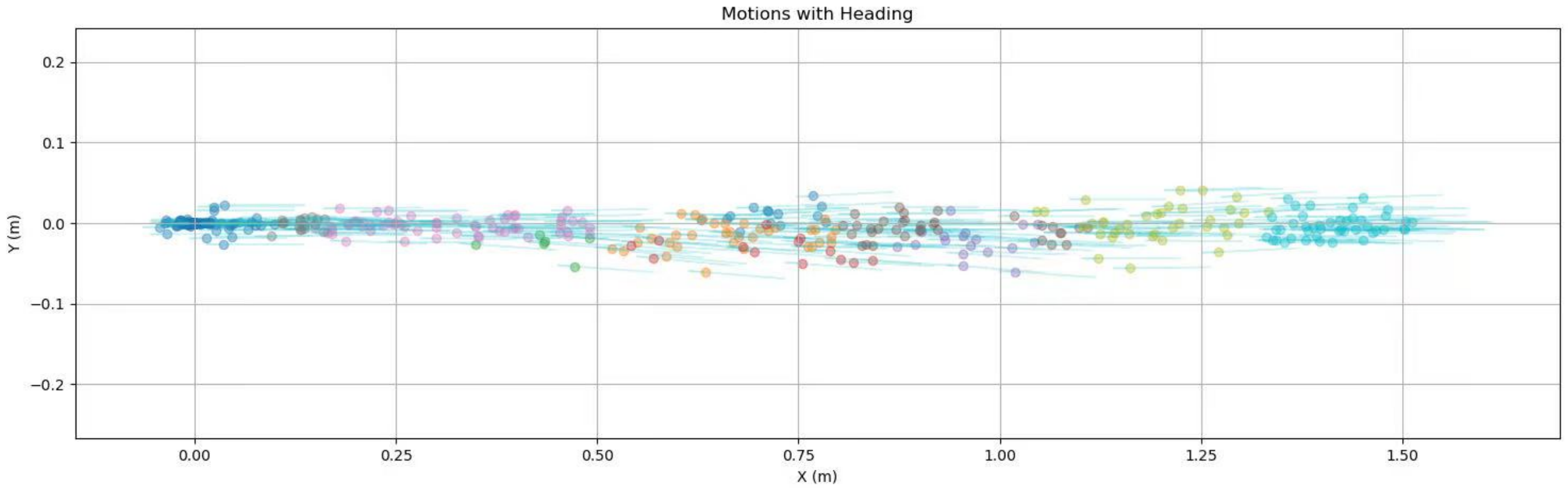}
  \caption{Visualization of clustered motion primitives.
  }\label{f5}
  \vspace{-10pt}
\end{figure*}

\subsection{Hybrid Trajectory Formulation and Visualization}
\label{secA3}

For trajectory prediction, we further design a hybrid formulation that combines discrete motion primitive \cite{zhou2025autovla} retrieval with continuous offset refinement, improving both the robustness and precision of trajectory planning. Specifically, we first uniformly sample approximately 120K trajectories from the NAVSIM dataset and augment each trajectory with an origin-aligned stationary pose. The pose sequence of each trajectory is then converted into a sequence of vehicle bounding-box polygons according to the vehicle geometry. Based on these polygonal trajectories, we construct a pairwise average Euclidean distance matrix to measure trajectory similarity. A tolerance parameter tol is further introduced to estimate local trajectory density. We subsequently adopt a density-peak-based greedy clustering strategy to iteratively select representative trajectories as cluster centers while removing neighboring samples within the corresponding density region. The process continues until 2048 high-density trajectory centers are obtained or the density threshold reaches its lower bound. Finally, the pose sequences corresponding to these cluster centers are fixed as motion primitives, serving as semantic trajectory atoms for downstream trajectory generation. As illustrated in Figure \ref{f5}, we visualize the clustered motion primitives used for trajectory generation.
\begin{figure*}[!t]
  \centering
  \includegraphics[width=1.0\columnwidth]{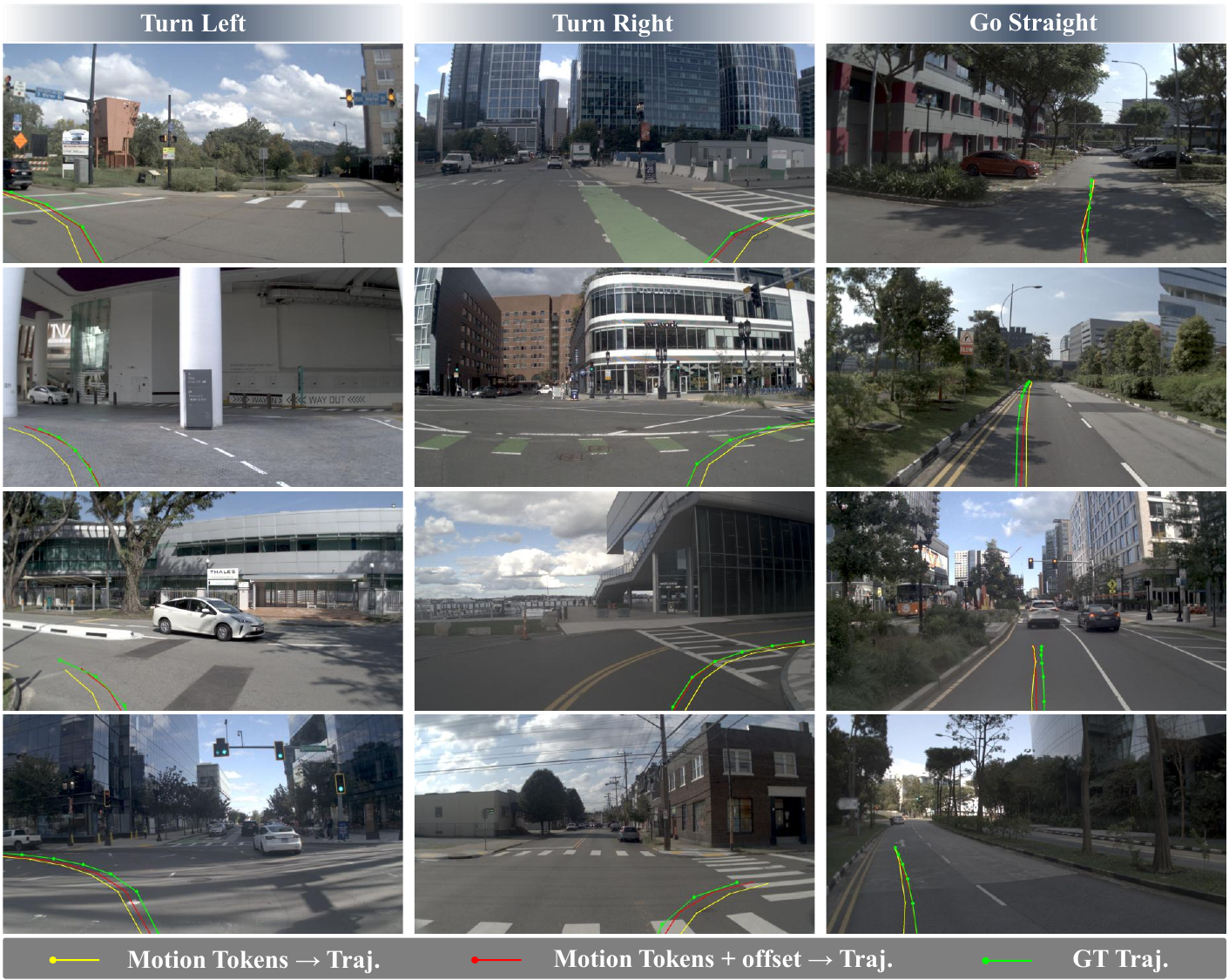}
  \caption{Visualization of the proposed hybrid trajectory formulation. Green denotes the human-driving ground-truth trajectory, yellow denotes the trajectory decoded solely from discrete motion tokens, and red denotes the final prediction after continuous offset refinement.
  }\label{f6}
\end{figure*}

During training, continuous trajectories are reformulated as combinations of discrete motion primitives and continuous offsets for trajectory supervision. Specifically, for each action timestep, a motion primitive is selected from the top-k nearest candidates ranked by trajectory similarity, where $k=10$ is empirically found to achieve the best performance. Random sampling is further applied within the candidate set to increase multimodal coverage among highly similar local trajectories. After matching the discrete motion primitives to the ground-truth trajectory rollout, the residual errors between the selected primitive and the target trajectory are explicitly computed, including positional and heading deviations. 
These continuous residuals are then modeled as offset refinement terms to compensate for the information loss introduced by trajectory discretization. As shown in Table \ref{tab-5}, the experimental results demonstrate that discrete motion primitives improve trajectory robustness, while continuous offset refinement recovers fine-grained prediction precision. Overall, the proposed hybrid formulation effectively balances robustness and accuracy, leading to substantial improvements in trajectory prediction performance.

Figure \ref{f6} visualizes the planning performance of the proposed hybrid trajectory formulation in three representative driving scenarios: going straight, turning left, and turning right. The green trajectory denotes the human-driving ground-truth trajectory, the yellow trajectory represents the prediction decoded solely from discrete motion primitives, and the red trajectory corresponds to the final prediction after continuous offset refinement. As shown in the figure, using only discrete motion primitives often leads to under-steering behavior in turning scenarios and insufficient trajectory smoothness in straight-driving cases, resulting in unstable steering adjustments. In contrast, continuous offset refinement effectively improves trajectory quality, leading to smoother, safer, and more accurate driving behaviors.

\begin{figure*}[!t]
  \centering
  \includegraphics[width=1.0\columnwidth]{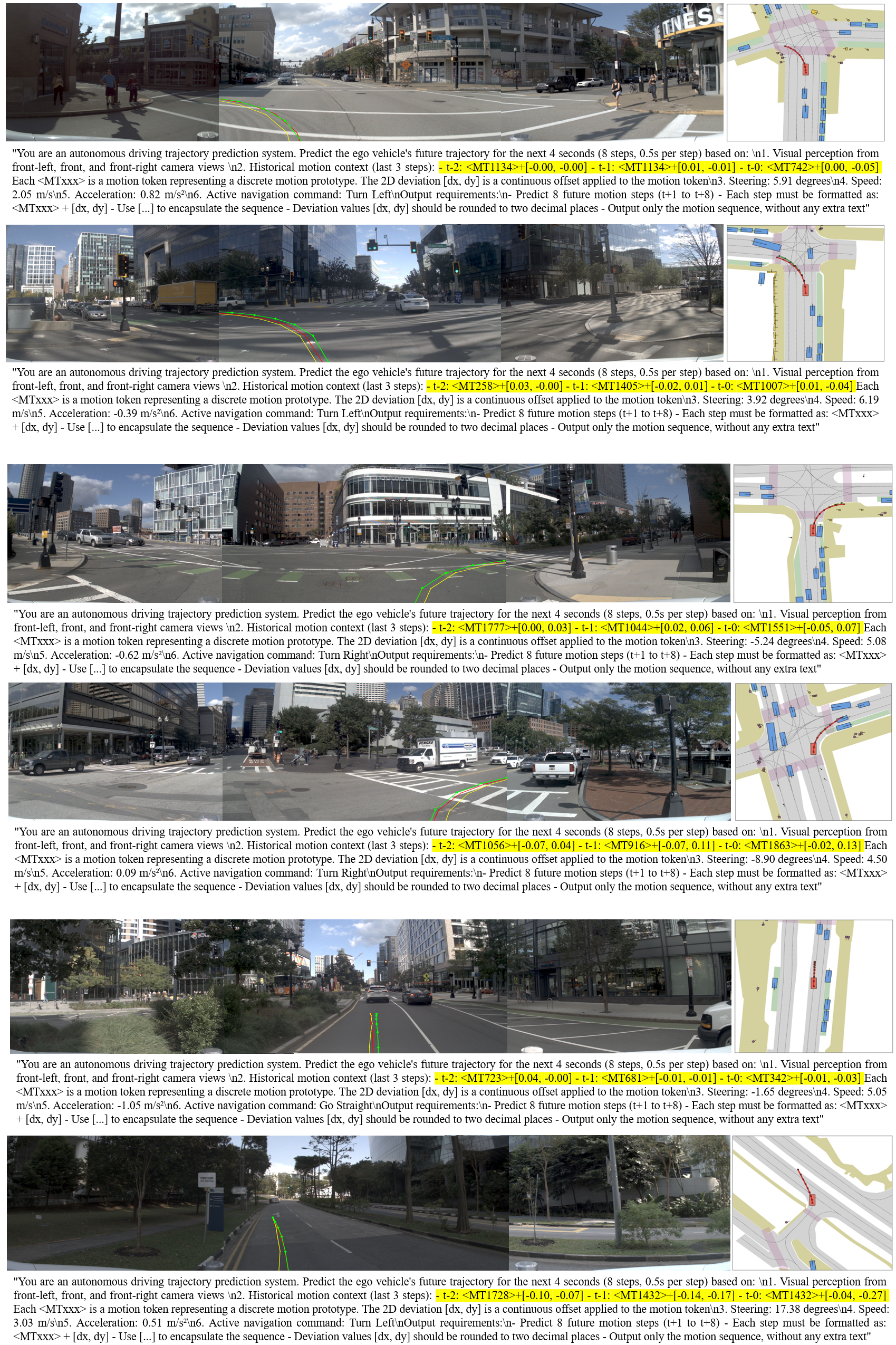}
  \caption{Visualization of the proposed hybrid trajectory formulation in representative autonomous driving scenarios, including left turns, right turns, and straight driving. 
  }\label{f7}
\end{figure*}

As illustrated in Figure \ref{f7}, additional visualizations of closed-loop trajectory planning are further provided. Specifically, three trajectories are visualized in the front-view image to provide an intuitive comparison. In addition, the instruction details used during training on the NAVSIM dataset are presented. The instruction includes the ego-vehicle state information, driving command, and 2-second historical trajectory. The historical trajectory is reconstructed as a combination of discrete motion primitive retrieval and continuous offset refinement  ($k = 1$ during reconstruction).

\end{document}